\documentclass[3p,times]{elsarticle}

\usepackage{ecrc}
\usepackage{hyperref}

\volume{00}

\firstpage{1}

\runauth{A. Gomez-Villa et al.}

\jid{phpro}

\jnltitlelogo{Vision Research}

\usepackage{amssymb}

\usepackage[figuresright]{rotating}
\usepackage{multicol,caption,graphicx,xcolor}
\usepackage{flushend}
\usepackage{balance}

\newenvironment{Figure}
  {\par\medskip\noindent\minipage{\linewidth}}
  {\endminipage\par\medskip}

\usepackage[all]{xy}
\usepackage{amsmath,amssymb}
\usepackage{mathrsfs}
\usepackage{tikz}

\usepackage[T1]{fontenc}
\usepackage[utf8]{inputenc}
\usepackage{tabularx,ragged2e,booktabs,caption}
\newcolumntype{C}[1]{>{\Centering}m{#1}}

\DeclareMathOperator*{\argminA}{arg\,min}

\begin{document}

\begin{frontmatter}



\dochead{}

\title{Visual Illusions Also Deceive Convolutional Neural Networks: Analysis and Implications}

\author[label1]{A. Gomez-Villa}
\ead{alexander.gomez@upf.edu}
\author[label1]{A. Mart\'in}
\ead{adrian.martin@upf.edu}
\author[label1]{J. Vazquez-Corral}
\ead{javier.vazquez@upf.edu}
\author[label1]{M. Bertalm\'{i}o}
\ead{marcelo.bertalmio@upf.edu}
\author[label2]{J. Malo}
\ead{jesus.malo@uv.es}
\address[label1]{Dept. Inf. Comm. Tech., Universitat Pompeu Fabra, Barcelona, Spain}
\address[label2]{Image Proc. Lab, Universitat de Val\`encia, Val\`encia, Spain}

\begin{abstract}
Visual illusions allow researchers to devise and test new models of visual perception. Here we show that artificial neural networks trained for basic visual tasks in natural images are deceived by brightness and color illusions, having a  response that is qualitatively very similar to the human achromatic and chromatic contrast sensitivity functions, and consistent with natural image statistics. We also show that, while these artificial networks are deceived by illusions, their response might be significantly different to that of humans. Our results suggest that low-level illusions appear in any system that has to perform basic visual tasks in natural environments, in line with error minimization explanations of visual function, and they also imply a word of caution on using artificial networks to study human vision, as previously suggested in other contexts in the vision science literature.
\end{abstract}


\end{frontmatter}


\begin{multicols}{2}

\section{Introduction}
\label{intro}

A visual illusion (VI) is an image stimulus that induces a visual percept that is not consistent with the visual information that can be physically measured in the scene. An example VI can be seen in Fig. \ref{fig:canon}: the center squares have the exact same gray value, and therefore send the same light intensity to our eyes (as a measurement with a photometer could attest), but we perceive the gray square over the white background as being darker than the gray square over the black background.
There are many types of VIs, involving for instance the perception of brightness \cite{white1979new,mccourt1982spatial,devalois1990spatial}, color \cite{Kitaoka05,Zaidi12,Loomis72,Brainard05,Abrams07}, texture \cite{Blakemore69,Ross91,Foley97,Watson97}, motion \cite{Solomon06,Mather08,Morgan11}, geometry \cite{Weintraub1971,Westheimer2008}, etc.

\begin{Figure}
\begin{center}
\hspace{-0.5cm}\includegraphics[width=0.5\linewidth]{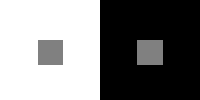}
\end{center}
\captionof{figure}{An example visual illusion. The squares have the same gray value, but one is perceived as being brighter than the other.}
\label{fig:canon}
\end{Figure}

In the context of the efficient representation view of biological vision \cite{Attneave1954,Barlow1961}
- which states that the organization of the visual system in general and neural responses in particular are tailored to the statistics of the images that the individual typically encounters, so that visual information can be encoded in the most efficient way-
VIs are not seen as failures but as a by-product of strategies to adapt to natural image statistics \cite{Barlow90,Clifford00,Clifford02,Cliford07,Laparra15}.
Therefore, VIs provide compelling case examples which are useful to probe theories about how our perception works.

Following classical works in visual neuroscience and visual perception \cite{Hubel1959,Campbell1968}
that successfully predicted visual responses as a linear filtering operation followed by a pointwise nonlinearity, 
the ``standard model'' of vision \cite{Olshausen2005} has become that
of a filter bank or rather a cascade of linear and nonlinear (L+NL) modules \cite{Carandini12,Martinez18}.
In computer vision, Artificial Neural Networks (ANNs) have shown outstanding achievements for many applications, where they are able to match human performance in tasks like face recognition and object classification.
The design of ANNs
has been motivated by analogy with the brain, with neurobiological models as the source of inspiration  \cite{Haykin2009}, and for this reason ANNs can be seen as constituted by linear and nonlinear (L+NL) modules as well.

But despite the fact that ANNs are inspired by biological networks, they fail to emulate basic perceptual phenomena.
For example, ANNs are prone to adversarial attacks,
where a very small change in pixel values in an image of some object A can lead the neural network to misclassify it as being a picture of object B, while for a human observer both the original and the modified images are perceived as being identical \cite{Goodfellow2018}.
Many attacks have been demonstrated, and 
few strong countermeasures
exist for them \cite{Goodfellow2018};
part of the problem is that the decisions of ANNs can't be interpreted, the network acts as a black box, even to its designers \cite{Jaspers2019}.
Another example is that the classification performance of ANNs falls rapidly when noise or texture changes are introduced on the test images, while human performance remains fairly stable under these modifications \cite{Geirhos2018}.
These failures highlight another key limitation of ANNs: they require hundreds of thousands times more information than humans in order to achieve human-level performance in a given task \cite{inverse}. 
To all this we may add that there is a major quantitative difference between ANNs and vision models: artificial networks are usually trained for visual tasks related to image processing and computer vision,  whereas  models in vision science are developed to reproduce a range of psychophysical or physiological phenomena. As a result, while artificial networks may excel in the specific goals they were optimized for, they may miss basic psychophysical facts \cite{Martinez19,Wichmann19}.

Since 2018, a handful of works 
have found that convolutional neural networks (CNNs) trained in natural images can also be ``fooled'' by VIs, in the sense that their response to an image input that is a VI for a human is (qualitatively) the same as that of humans, and therefore inconsistent with the actual physical values of the light stimulus. This has been shown for VIs of very different type: motion \cite{watanabe2018illusory}, brightness and color \cite{GomezCVPR19}, completion \cite{kim2019neural}, or geometry \cite{Ward19}.
This 
very recent line of research, devoted to the study of similarities and differences between the VIs suffered by human viewers and artificial neural networks, may be relevant
to explore the limitations of simplified architectures and suggest better models of biological vision.


The current work expands our initial findings on visual illusions that deceive artificial neural networks \cite{GomezCVPR19}. Our contributions in this paper are:
\begin{enumerate}
\item Providing more exhaustive confirmation that CNNs trained for basic visual tasks in natural images are deceived by brightness and color illusions in complex spatial contexts.

\item Showing, based on a linear approximation analysis of CNNs, that these architectures have a response that is qualitatively very similar to the human achromatic and chromatic contrast sensitivity functions (CSFs), and consistent with natural image statistics.

\item Performing a psychophysical-like analysis of CNNs to show that, while these artificial networks are deceived by illusions, their nature might be significantly different to that of humans.

\end{enumerate}

These contributions suggest the following.

From result (1) above, low-level VIs may appear in any system that has to perform basic visual tasks in natural environments.

From (2), and in line with
error minimization explanations of visual 
function \cite{Atick93,MacLeod01,MacLeod03,Laparra12,Laparra15},
CNNs also develop achromatic and opponent chromatic channels with band-pass/low-pass spatial frequency response because the
optimal removal of non-natural features leads to the identification of principal directions in the image statistics of natural scenes. 

More interestingly, from (3), discrepancies with humans in quantitative experiments imply a word of caution on using CNNs to study human vision, as previously suggested in other contexts (with regards to L+NL formulations) in the vision science literature \cite{Wandell1995,Carandini2005,Olshausen2013}.

\begin{figure*}[t]
\begin{center}
\includegraphics[width=0.8\linewidth]{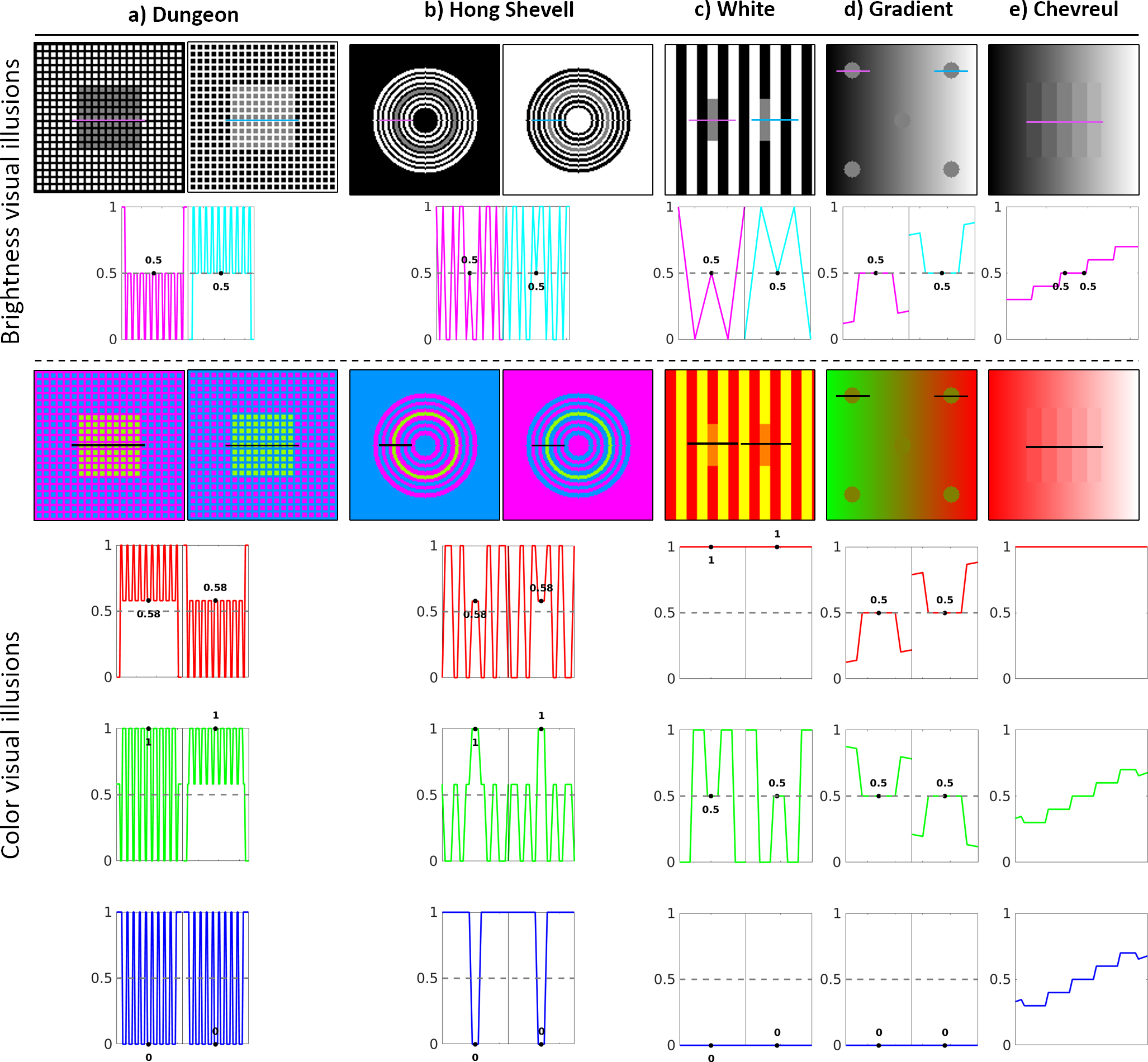}
\end{center}
\caption{Stimuli to check the existence of brightness and chromatic visual illusions. Profiles represent the RGB digital values of the stimuli at the lines depicted in the images. Equal digital values imply physically equal tests. Perception is very different though.}
\label{fig_stimuli1}
\end{figure*}

The structure of the paper is as follows. In the \emph{Materials and Methods} section we introduce the stimuli used in the experiments, we describe the considered architectures and the visual tasks used to train them, and we discuss two alternative methods to describe the illusions that may be found in the networks.
In the \emph{Results} section we compute the shifts of the responses due to context and the corresponding pairs of the networks in asymmetric matching experiments. In the \emph{Linear Analysis} section, eigenanalysis of the networks reveals intrinsic filters which are similar to the CSFs in opponent channels, and finally, the \emph{Discussion} analyzes the implications of the results in terms of complexity of the networks and appropriateness to model human vision.

\section{Materials and Methods}
\label{methods}

\subsection{The stimuli}

In this work we deal with two sets of stimuli. 
First, classical tests shown in Fig.~\ref{fig_stimuli1} are used to point out that convolutional networks have illusions that are \emph{reasonably} similar to human viewers. The classical visual illusions in Fig.~\ref{fig_stimuli1} present targets that are physically the same but are seen differently depending on their surrounds, with induction perceptual phenomena of {\it assimilation} (when the perception of the target shifts towards that of its surround) or {\it contrast} (when the perception of the target moves away from that of its surround).
The targets are, in the Dungeon illusion \cite{bressan2001explaining}, Fig. \ref{fig_stimuli1}a, the large central squares, in Hong-Shevell \cite{hong2004brightness}, Fig. \ref{fig_stimuli1}b,  the middle rings, in the White illusion \cite{white1979new}, Fig. \ref{fig_stimuli1}c,  the small grey bars, and in the Luminance gradient illusion (combination of \cite{bruke,adelson2000zj}, Fig. \ref{fig_stimuli1}d,  the circles.
The fact that the targets are identical can be seen in the second and fourth to sixth rows of Fig. \ref{fig_stimuli1}, that plots the digital image values along some line segments shown over the visual illusions in the first and third rows.
The Chevreul illusion  ~\cite{ratliff1965mach}, Fig. \ref{fig_stimuli1}e, presents homogeneous bands of increasing intensity, from left to right, but these bands are perceived to be in-homogeneous, with darker and brighter lines at the borders between adjacent bands.

Then, a second experiment simulating asymmetric color matching is performed with the networks to have results that are quantitatively comparable to those of human viewers. 
In a color matching experiment, given an image stimulus consisting of a test patch of color {\bf t} over an inducing surround {\bf s},
the \emph{observer} looks for the \emph{corresponding pair} color {\bf t'} that seen on a neutral reference background matches the perception of the test {\bf t}.
In this case we study the behavior of the network in the corresponding pair experiments using different inductors {\bf s} in the setting described in \cite{ware82}, see the chromatic content of the scenes in Fig.~\ref{datos_wc} (with tests/inductors of 30 $cd/m^2$). 

\begin{Figure}
\begin{center}
\hspace{-0.5cm}\includegraphics[width=0.8\linewidth]{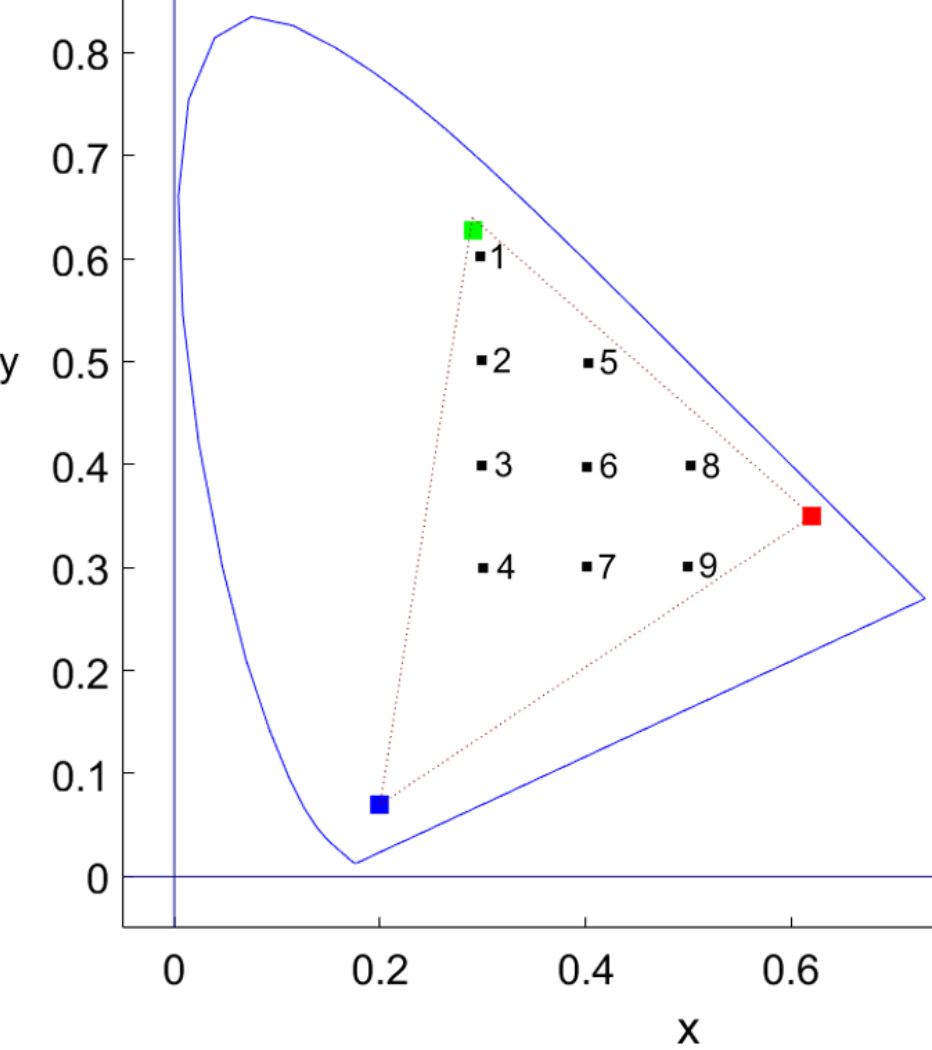}
\end{center}
\captionof{figure}{Test points and inductors in Ware-Cowan corresponding pair experiment \cite{ware82}.}
\label{datos_wc}
\end{Figure}

\subsection{The networks: architectures and training}

We trained two CNN architectures for three low-level visual tasks: \emph{denoising} and \emph{deblurring} (as in the work of Gomez-Villa et al.~\cite{GomezCVPR19})
and also \emph{restoration} (a mixture of the denoising and deblurring problems). Hence, we have 6 models. 
The general setting to obtain the parameters of the models is supervised learning (see Fig.~\ref{network}).

For consistency with the spatial extent of the stimuli used in the experiment with humans reported by Ware-Cowan~\cite{ware82}, 
we assume the images subtend 1.83 deg with sampling frequency of 70 cpd ($128\times128$ pixels).

The first architecture has input and output layers of size $128\times128\times3$ pixels.  
The architecture has \emph{two} hidden layers with \emph{eight} feature maps with a kernel size of $5\times5$ and no stride, and sigmoid activation functions. 
The second architecture is a bit deeper and hence with substantially more free parameters. It also has input and output layers of size $128\times128\times3$, but \emph{four} hidden layers with $24$ feature maps. Kernel sizes and non-linearities are the same as in the first architecture. 
The two hidden-layer architectures (shallow) are named with respect to the task they are trained for: DN-NET (denoising network), DB-NET (deblurring network), and RestoreNet (restauration network). As for the four hidden-layer architectures (deep), we added the "Deep" word to the corresponding shallow architecture name, hence: Deep DN-NET, Deep DB-NET, and Deep RestoreNet.

\begin{Figure}
\begin{center}
\hspace{-0.5cm}\includegraphics[width=0.9\linewidth]{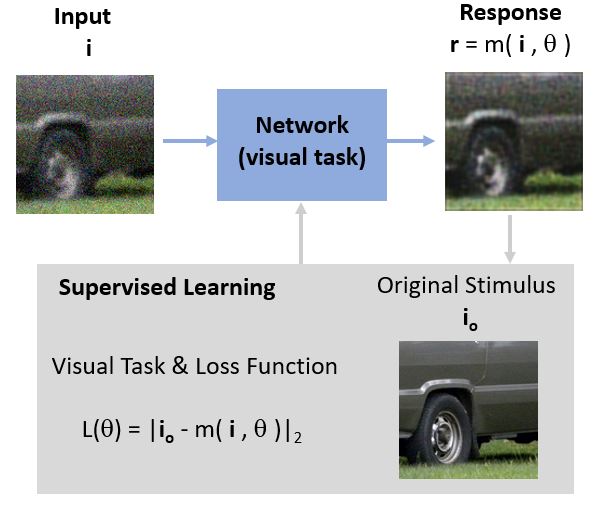}
\end{center}
\captionof{figure}{The convolutional architectures considered here take regular photographic images as input stimuli and perform some visual task (e.g. denoising, deblurring, restoration, etc.) after training through supervised learning.}
\label{network}
\end{Figure}

 Mean squared error was used as loss function in all the tasks and all the models were implemented\footnote{source code will be made publicly available} using Tensorflow~\cite{tensorflow}. The maximum number of epochs was set to 100 and we early stop the optimization if there is no improvement in the validation set after 5 consecutive evaluations. 

The dataset used for training all the architectures is the Large Scale Visual Recognition Challenge 2014 CLS-LOC validation dataset (which contains 50k images), leaving 10k images for validation purposes. This dataset is a subset of the whole ImageNet dataset ~\cite{russakovsky2015imagenet}.

For denoising, we corrupted the images with additive Gaussian noise of $\sigma = 25$ in each RGB channel (digital counts in the range [0, 255]). In the case of deblurring, we blurred the images with a spatial Gaussian kernel of width $\sigma_x = 0.03$ deg (2 pixels). As for restoration, first we blurred the images with a Gaussian kernel of $\sigma_x = 0.03$ deg and then we corrupted the images with additive Gaussian noise of $\sigma = 25$. 

Note that the restoration task combines the other two tasks, thus being more general. 

The above case-study architectures with 2 and 4 hidden layers give us the opportunity to check the behavior of the networks in a systematic and comparable way. However, it is also interesting to explore the eventual illusions happening in current much deeper networks used in real image processing applications. To this end, we also explored the behavior of the modern 17 layer denoising CNN  by Zhang. et al.~\cite{zhang2017beyond} using a trained implementation provided by the authors\footnote{See \url{https://github.com/cszn/DnCNN}.  }.

\subsection{Strategies to describe the illusions in ANNs}
In this work we use two alternative strategies to describe the visual illusions of the network.
The first strategy is inspired in physiology while the second simulates psychophysics.

The physiological-like strategy consists of measuring the shifts in the responses of the units of the network tuned to the location of the test.
In principle, this could be done at the final layer of the network or at any intermediate hidden layer.
In any case, given the remarkable differences between artificial neurons and actual neurons, quantitative comparison of these shifts with human perception is not obvious. 
Moreover, the analysis of the deviations of the responses implicitly assumes certain \emph{interpretation} of the responses by the network.
For instance, the considered CNNs lead to an output response which is in the same domain as the input. In cases like this, it is reasonable to describe the change in \emph{perceived color} by analyzing the changes in the response in the last layer as if it was a regular image. 
This definition of perceived color is also done in certain vision models~\cite{Otazu08,otazu2010toward}, where the perceptions are described from the linearly reconstructed image that comes from an inner representation which is nonlinear.
In summary, it is important to acknowledge that this definition of \emph{perceived color} makes the extra assumption that the model interprets the responses in a particular way (e.g. in the input space). 
This assumption is relevant because, in general, the signal representation of other artificial nets may not be similar to the input domain. 

The psychophysical-like description does not make this (extra) assumption on the way the network interprets changes of the response. In this case, we simulate perceptual matches as it's done in psychophysics. Examples include asymmetric color matching experiments or corresponding pairs~\cite{heinemann1955,ware82}.
In this corresponding pair setting, the observer (or the network) changes the stimulus in the input space in a context-free reference until the inner response (whatever this response is) equals the response obtained from the test with context. 
In this situation, matching the appearance means getting the same response, and no assumption on how this response is interpreted has to be done. 

The description based on shifts of the responses suggest trends in the behavior of the network but it is not easily comparable to human performance.
On the contrary, in simulating asymmetric matching, the units to quantify the strength of the illusion of the network are exactly the same as in human psychophysics. Therefore, the performance of CNNs and humans is completely comparable in this second case.


\section{Results}

We did two numerical experiments with the  \emph{shallow} and the \emph{deep} CNNs trained according to the low-level visual tasks considered above.  

In these experiments where the networks are applied to the illusion-inducing stimuli, it is convenient to think on stimuli, the images, $\mathbf{i}$, and responses, $\mathbf{r}$, as column vectors where \emph{test} and \emph{surround} are spatially disjoint, i.e. in different rows of the corresponding column vector. Schematically,
\begin{equation}
\mathbf{i} = \left[\!\!
\begin{array}{c}
\mathbf{t}\\
\mathbf{s}\\
\end{array}
\!\!\right]
\,\,\,\,\,\,\textrm{and}\,\,\,\,\,\,
\mathbf{r}=\left[\!\!
\begin{array}{c}
\mathbf{r}_t\\
\mathbf{r}_s\\
\end{array}
\!\!\right]
\end{equation}

The first experiment consists of computing the response of the network for the physically identical tests seen in different spatio-chromatic contexts (the stimuli in Fig.~\ref{fig_stimuli1}).
This experiment reduces to applying the $k-$th model to each stimulus, $\mathbf{i}$, to compute the corresponding response, $\mathbf{r}$:
\begin{equation}
    \mathbf{r} = m(\mathbf{i},\theta_k)
    \label{response}
\end{equation}
where $\theta_k$ stands for the parameters learnt with certain architecture and task. With 2 architectures and 3 tasks, $k = 1,\ldots,6$. Also, we add a recent very deep CNN devised for image denoising to these experiments \cite{zhang2017beyond}. 

The second experiment consists of obtaining the corresponding pairs of the test colors represented by black squares in Fig.~\ref{datos_wc}, seen in backgrounds defined by the R, G and B inductors (color triangles in Fig.~\ref{datos_wc}).
Given that the networks were trained from images expressed in digital counts we did not considered all the test colors used in~\cite{ware82}, but only those within the triangle of colors available from RGB digital counts. For the same reason the inductors used in our experiment are slightly different from those used in the original experiment, but, as disscused later, this will not be a major problem in the interpretation of the results.

\subsection{Shifts in the response}

The qualitative interpretation of the behavior of the responses in different settings can be done by checking if the response shifts in certain direction. For instance, in the achromatic cases, \emph{is the response departing from average brightness?}. If yes, \emph{in which direction, darker or brighter?}. In the color cases, the change of perceived hue question could be for instance \emph{are the values of the response departing from green?}, if yes, \emph{in which direction, towards red-yellow?} 

\subsubsection{Achromatic case}
 
 The second row of Figure \ref{fig_stimuli1} shows that the digital value (and hence the luminance) of the tests in the considered stimuli is the same, however, the brightness we perceive is markedly different (see the images in the first row of Figure \ref{fig_stimuli1}).
  In particular, in the case of Dungeon, Hong-Shevell and White illusions we perceive the left target to be darker than the right one. Conversely, in the Gradient illusion the effect is the opposite, the left target seems darker than the right one. In Chevreul, each band is perceived to be lighter and darker in its left and right sides respectively.
 
\begin{figure*}
\begin{center}
\hspace{-0.5cm}\includegraphics[width=0.75\linewidth]{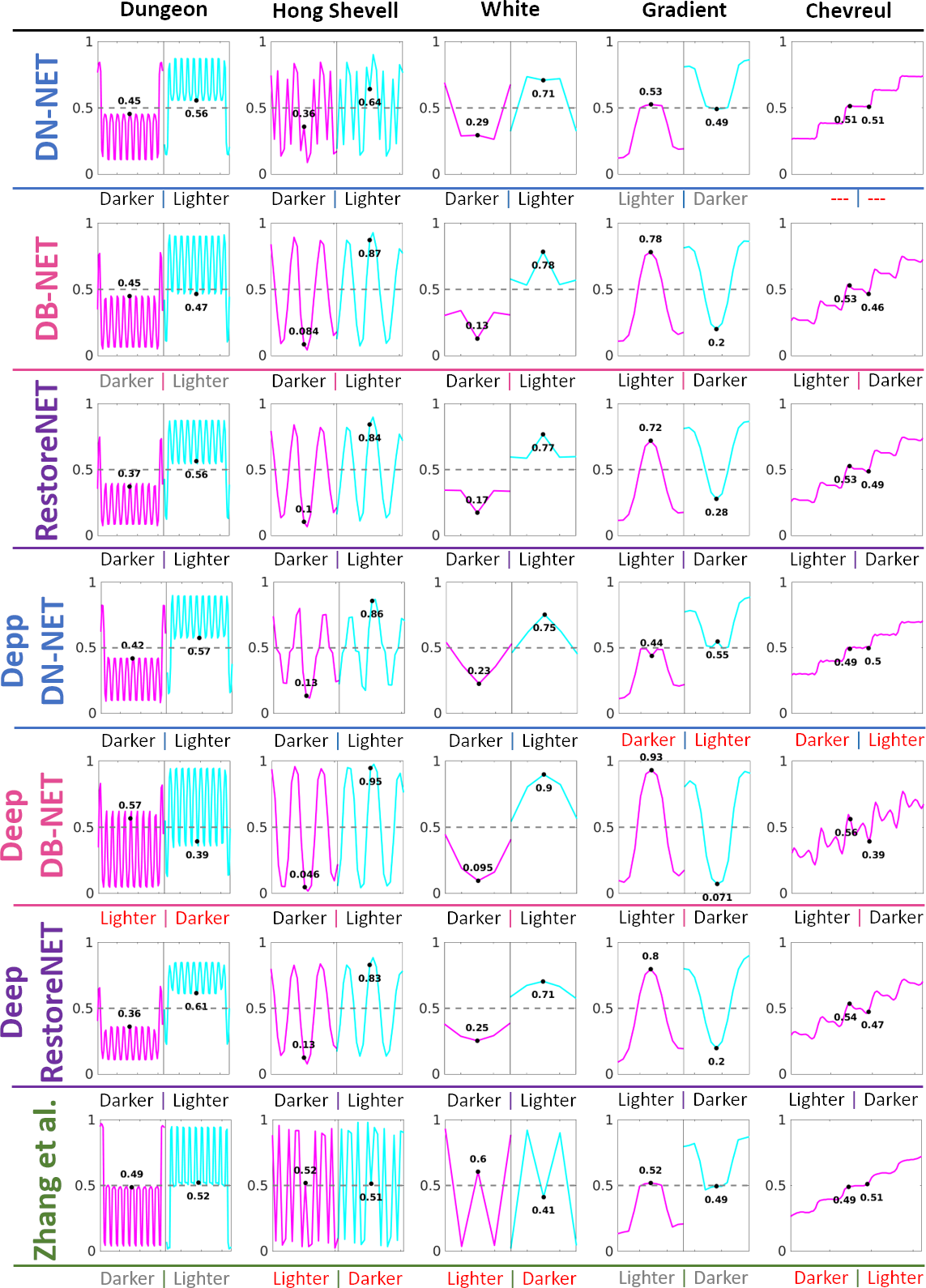}
\end{center}
\captionof{figure}{Response of the different CNNs to the stimuli inducing brightness illusions. Below the response profile in each case we describe the direction of the shift (darker or lighter). Descriptions in \emph{black} indicate the shift is in the same direction as humans. Descriptions in \emph{gray} also mean correspondence with humans but weak effect. And those in \emph{red} mean non-human shifts.}
\label{fig:grayVI}
\end{figure*}

 Figure \ref{fig:grayVI} shows the responses' profiles obtained from the different CNN considered in this work when they are fed with the achromatic illusion-inducing images. We can see that the shallow CNNs trained for denoising (DN-NET), deblurring (DB-NET) and restoration -i.e. denoising and deblurring at the same time- (RestoreNET) replicate the four first illusions, i.e. they show a response variation in the same direction than human perception. Additionally, DB-NET and RestoreNET also replicate the Chevreul illusion. If we move to looking at the deeeper CNNs, the results vary much more. In this case, only the CNN trained for restoration (Deep RestoreNET) shows a response that matches human perception for the five illusions considered. 
  The denoising CNN (Deep DN-NET) is not able to reproduce the Gradient and the Chevreul illusions, while the deblurring CNN (Deep DB-NET) fails at matching human perception in the Dungeon illusion.
 
 Finally, the modern -and really deep- CNN from Zhang \emph{et al.} \cite{zhang2017beyond} does not modify much the input images. Therefore, it fails to reproduce three out of five cases. And in the Dungeon and Gradient cases in which the direction of change is human-like, the magnitude of the change is really small.

In summary, simpler networks have shifts in perceived brightness in similar directions as humans (in about 80\% of the cases). On the contrary, a really deep network trained for the same kind of goal has very small brightness illusions, and in most cases, not in the human direction.

\begin{figure*}
\begin{center}
\includegraphics[width=0.75\linewidth]{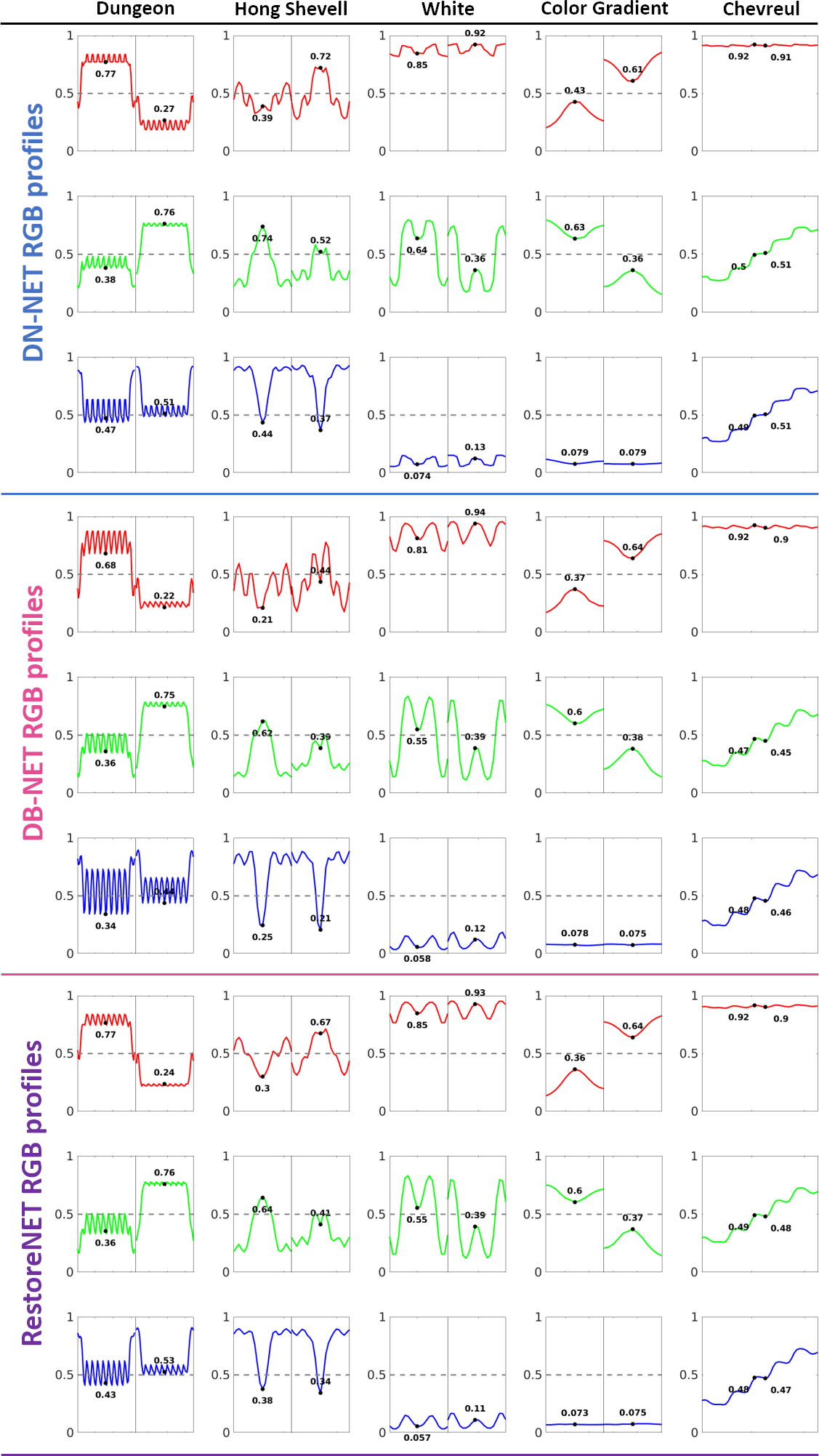}
\end{center}
\captionof{figure}{Response profiles of the shallow CNNs to the color stimuli that illustrate different color illusions. The responses of the sensors tuned to RGB hues at the highlighted locations represent the \emph{perception} of the network for the considered tests.}
\label{fig:colorDNNET}
\end{figure*}

\begin{table*}[]
\footnotesize
\begin{tabular}{cccccccccccccccc}
 & \multicolumn{15}{c}{DN-NET} \\ \cline{2-16} 
\multicolumn{1}{c|}{} & \multicolumn{3}{c|}{Dungeon} & \multicolumn{3}{c|}{Hong-Shevell} & \multicolumn{3}{c|}{White} & \multicolumn{3}{c|}{Gradient} & \multicolumn{3}{c|}{Chevreul} \\ \cline{2-16} 
\multicolumn{1}{c|}{} & \multicolumn{1}{c|}{In} & Out-L & \multicolumn{1}{c|}{Out-R} & \multicolumn{1}{c|}{In} & Out-L & \multicolumn{1}{c|}{Out-R} & \multicolumn{1}{c|}{In} & Out-L & \multicolumn{1}{c|}{Out-R} & \multicolumn{1}{c|}{In} & Out-L & \multicolumn{1}{c|}{Out-R} & \multicolumn{1}{c|}{In} & Out-L & \multicolumn{1}{c|}{Out-R} \\ \hline
\multicolumn{1}{|c|}{R} & \multicolumn{1}{c|}{0.58} & 0.77 & \multicolumn{1}{c|}{0.27} & \multicolumn{1}{c|}{0.58} & 0.39 & \multicolumn{1}{c|}{0.72} & \multicolumn{1}{c|}{1} & 0.85 & \multicolumn{1}{c|}{0.92} & \multicolumn{1}{c|}{0.5} & 0.43 & \multicolumn{1}{c|}{0.61} & \multicolumn{1}{c|}{1} & 0.92 & \multicolumn{1}{c|}{0.91} \\
\multicolumn{1}{|c|}{G} & \multicolumn{1}{c|}{1} & 0.38 & \multicolumn{1}{c|}{0.76} & \multicolumn{1}{c|}{1} & 0.74 & \multicolumn{1}{c|}{0.53} & \multicolumn{1}{c|}{0.5} & 0.64 & \multicolumn{1}{c|}{0.36} & \multicolumn{1}{c|}{0.5} & 0.63 & \multicolumn{1}{c|}{0.36} & \multicolumn{1}{c|}{0.5} & 0.5 & \multicolumn{1}{c|}{0.51} \\
\multicolumn{1}{|c|}{B} & \multicolumn{1}{c|}{0} & 0.47 & \multicolumn{1}{c|}{0.51} & \multicolumn{1}{c|}{0} & 0.44 & \multicolumn{1}{c|}{0.37} & \multicolumn{1}{c|}{0} & 0.074 & \multicolumn{1}{c|}{0.13} & \multicolumn{1}{c|}{0} & 0.079 & \multicolumn{1}{c|}{0.079} & \multicolumn{1}{c|}{0.5} & 0.49 & \multicolumn{1}{c|}{0.51} \\ \hline
 &  &  &  &  &  &  &  &  &  &  &  &  &  &  &  \\
 & \multicolumn{15}{c}{DB-NET} \\ \cline{2-16} 
\multicolumn{1}{c|}{} & \multicolumn{3}{c|}{Dungeon} & \multicolumn{3}{c|}{Hong-Shevell} & \multicolumn{3}{c|}{White} & \multicolumn{3}{c|}{Gradient} & \multicolumn{3}{c|}{Chevreul} \\ \cline{2-16} 
\multicolumn{1}{c|}{} & \multicolumn{1}{c|}{In} & Out-L & \multicolumn{1}{c|}{Out-R} & \multicolumn{1}{c|}{In} & Out-L & \multicolumn{1}{c|}{Out-R} & \multicolumn{1}{c|}{In} & Out-L & \multicolumn{1}{c|}{Out-R} & \multicolumn{1}{c|}{In} & Out-L & \multicolumn{1}{c|}{Out-R} & \multicolumn{1}{c|}{In} & Out-L & \multicolumn{1}{c|}{Out-R} \\ \hline
\multicolumn{1}{|c|}{R} & \multicolumn{1}{c|}{0.58} & 0.68 & \multicolumn{1}{c|}{0.22} & \multicolumn{1}{c|}{0.58} & 0.21 & \multicolumn{1}{c|}{0.44} & \multicolumn{1}{c|}{1} & 0.81 & \multicolumn{1}{c|}{0.94} & \multicolumn{1}{c|}{0.5} & 0.37 & \multicolumn{1}{c|}{0.64} & \multicolumn{1}{c|}{1} & 0.92 & \multicolumn{1}{c|}{0.9} \\
\multicolumn{1}{|c|}{G} & \multicolumn{1}{c|}{1} & 0.36 & \multicolumn{1}{c|}{0.75} & \multicolumn{1}{c|}{1} & 0.62 & \multicolumn{1}{c|}{0.39} & \multicolumn{1}{c|}{0.5} & 0.55 & \multicolumn{1}{c|}{0.39} & \multicolumn{1}{c|}{0.5} & 0.6 & \multicolumn{1}{c|}{0.38} & \multicolumn{1}{c|}{0.5} & 0.47 & \multicolumn{1}{c|}{0.45} \\
\multicolumn{1}{|c|}{B} & \multicolumn{1}{c|}{0} & 0.34 & \multicolumn{1}{c|}{0.44} & \multicolumn{1}{c|}{0} & 0.25 & \multicolumn{1}{c|}{0.21} & \multicolumn{1}{c|}{0} & 0.058 & \multicolumn{1}{c|}{0.12} & \multicolumn{1}{c|}{0} & 0.078 & \multicolumn{1}{c|}{0.075} & \multicolumn{1}{c|}{0.5} & 0.48 & \multicolumn{1}{c|}{0.46} \\ \hline
 &  &  &  &  &  &  &  &  &  &  &  &  &  &  &  \\
 & \multicolumn{15}{c}{RestoreNET} \\ \cline{2-16} 
\multicolumn{1}{c|}{} & \multicolumn{3}{c|}{Dungeon} & \multicolumn{3}{c|}{Hong-Shevell} & \multicolumn{3}{c|}{White} & \multicolumn{3}{c|}{Gradient} & \multicolumn{3}{c|}{Chevreul} \\ \cline{2-16} 
\multicolumn{1}{c|}{} & \multicolumn{1}{c|}{In} & Out-L & \multicolumn{1}{c|}{Out-R} & \multicolumn{1}{c|}{In} & Out-L & \multicolumn{1}{c|}{Out-R} & \multicolumn{1}{c|}{In} & Out-L & \multicolumn{1}{c|}{Out-R} & \multicolumn{1}{c|}{In} & Out-L & \multicolumn{1}{c|}{Out-R} & \multicolumn{1}{c|}{In} & Out-L & \multicolumn{1}{c|}{Out-R} \\ \hline
\multicolumn{1}{|c|}{R} & \multicolumn{1}{c|}{0.58} & 0.77 & \multicolumn{1}{c|}{0.24} & \multicolumn{1}{c|}{0.58} & 0.3 & \multicolumn{1}{c|}{0.67} & \multicolumn{1}{c|}{1} & 0.85 & \multicolumn{1}{c|}{0.93} & \multicolumn{1}{c|}{0.5} & 0.36 & \multicolumn{1}{c|}{0.64} & \multicolumn{1}{c|}{1} & 0.92 & \multicolumn{1}{c|}{0.9} \\
\multicolumn{1}{|c|}{G} & \multicolumn{1}{c|}{1} & 0.36 & \multicolumn{1}{c|}{0.76} & \multicolumn{1}{c|}{1} & 0.64 & \multicolumn{1}{c|}{0.41} & \multicolumn{1}{c|}{0.5} & 0.55 & \multicolumn{1}{c|}{0.39} & \multicolumn{1}{c|}{0.5} & 0.6 & \multicolumn{1}{c|}{0.37} & \multicolumn{1}{c|}{0.5} & 0.49 & \multicolumn{1}{c|}{0.48} \\
\multicolumn{1}{|c|}{B} & \multicolumn{1}{c|}{0} & 0.43 & \multicolumn{1}{c|}{0.53} & \multicolumn{1}{c|}{0} & 0.38 & \multicolumn{1}{c|}{0.34} & \multicolumn{1}{c|}{0} & 0.057 & \multicolumn{1}{c|}{0.11} & \multicolumn{1}{c|}{0} & 0.073 & \multicolumn{1}{c|}{0.075} & \multicolumn{1}{c|}{0.5} & 0.48 & \multicolumn{1}{c|}{0.47} \\ \hline
\end{tabular}\caption{Input and model responses for the shallow CNNs for the different visual illusions studied.}\label{Tableforcolorillusions}
\end{table*}

\begin{table*}[]
\footnotesize
\begin{tabular}{cccccccccccccccc}
 & \multicolumn{15}{c}{Deep DN-NET} \\ \cline{2-16} 
\multicolumn{1}{c|}{} & \multicolumn{3}{c|}{Dungeon} & \multicolumn{3}{c|}{Hong-Shevell} & \multicolumn{3}{c|}{White} & \multicolumn{3}{c|}{Gradient} & \multicolumn{3}{c|}{Chevreul} \\ \cline{2-16} 
\multicolumn{1}{c|}{} & \multicolumn{1}{c|}{In} & Out-L & \multicolumn{1}{c|}{Out-R} & \multicolumn{1}{c|}{In} & Out-L & \multicolumn{1}{c|}{Out-R} & \multicolumn{1}{c|}{In} & Out-L & \multicolumn{1}{c|}{Out-R} & \multicolumn{1}{c|}{In} & Out-L & \multicolumn{1}{c|}{Out-R} & \multicolumn{1}{c|}{In} & Out-L & \multicolumn{1}{c|}{Out-R} \\ \hline
\multicolumn{1}{|c|}{R} & \multicolumn{1}{c|}{0.58} & 0.77 & \multicolumn{1}{c|}{0.27} & \multicolumn{1}{c|}{0.58} & 0.42 & \multicolumn{1}{c|}{0.63} & \multicolumn{1}{c|}{1} & 0.94 & \multicolumn{1}{c|}{0.97} & \multicolumn{1}{c|}{0.5} & 0.47 & \multicolumn{1}{c|}{0.49} & \multicolumn{1}{c|}{1} & 0.96 & \multicolumn{1}{c|}{0.95} \\
\multicolumn{1}{|c|}{G} & \multicolumn{1}{c|}{1} & 0.38 & \multicolumn{1}{c|}{0.72} & \multicolumn{1}{c|}{1} & 0.57 & \multicolumn{1}{c|}{0.4} & \multicolumn{1}{c|}{0.5} & 0.61 & \multicolumn{1}{c|}{0.38} & \multicolumn{1}{c|}{0.5} & 0.54 & \multicolumn{1}{c|}{0.48} & \multicolumn{1}{c|}{0.5} & 0.49 & \multicolumn{1}{c|}{0.49} \\
\multicolumn{1}{|c|}{B} & \multicolumn{1}{c|}{0} & 0.5 & \multicolumn{1}{c|}{0.57} & \multicolumn{1}{c|}{0} & 0.56 & \multicolumn{1}{c|}{0.49} & \multicolumn{1}{c|}{0} & 0.028 & \multicolumn{1}{c|}{0.056} & \multicolumn{1}{c|}{0} & 0.027 & \multicolumn{1}{c|}{0.03} & \multicolumn{1}{c|}{0.5} & 0.52 & \multicolumn{1}{c|}{0.52} \\ \hline
 &  &  &  &  &  &  &  &  &  &  &  &  &  &  &  \\
 & \multicolumn{15}{c}{Deep DB-NET} \\ \cline{2-16} 
\multicolumn{1}{c|}{} & \multicolumn{3}{c|}{Dungeon} & \multicolumn{3}{c|}{Hong-Shevell} & \multicolumn{3}{c|}{White} & \multicolumn{3}{c|}{Gradient} & \multicolumn{3}{c|}{Chevreul} \\ \cline{2-16} 
\multicolumn{1}{c|}{} & \multicolumn{1}{c|}{In} & Out-L & \multicolumn{1}{c|}{Out-R} & \multicolumn{1}{c|}{In} & Out-L & \multicolumn{1}{c|}{Out-R} & \multicolumn{1}{c|}{In} & Out-L & \multicolumn{1}{c|}{Out-R} & \multicolumn{1}{c|}{In} & Out-L & \multicolumn{1}{c|}{Out-R} & \multicolumn{1}{c|}{In} & Out-L & \multicolumn{1}{c|}{Out-R} \\ \hline
\multicolumn{1}{|c|}{R} & \multicolumn{1}{c|}{0.58} & 0.67 & \multicolumn{1}{c|}{0.29} & \multicolumn{1}{c|}{0.58} & 0.36 & \multicolumn{1}{c|}{0.45} & \multicolumn{1}{c|}{1} & 0.89 & \multicolumn{1}{c|}{0.97} & \multicolumn{1}{c|}{0.5} & 0.5 & \multicolumn{1}{c|}{0.45} & \multicolumn{1}{c|}{1} & 0.95 & \multicolumn{1}{c|}{0.92} \\
\multicolumn{1}{|c|}{G} & \multicolumn{1}{c|}{1} & 0.27 & \multicolumn{1}{c|}{0.72} & \multicolumn{1}{c|}{1} & 0.51 & \multicolumn{1}{c|}{0.25} & \multicolumn{1}{c|}{0.5} & 0.48 & \multicolumn{1}{c|}{0.47} & \multicolumn{1}{c|}{0.5} & 0.56 & \multicolumn{1}{c|}{0.41} & \multicolumn{1}{c|}{0.5} & 0.52 & \multicolumn{1}{c|}{0.45} \\
\multicolumn{1}{|c|}{B} & \multicolumn{1}{c|}{0} & 0.41 & \multicolumn{1}{c|}{0.57} & \multicolumn{1}{c|}{0} & 0.52 & \multicolumn{1}{c|}{0.36} & \multicolumn{1}{c|}{0} & 0.024 & \multicolumn{1}{c|}{0.094} & \multicolumn{1}{c|}{0} & 0.068 & \multicolumn{1}{c|}{0.052} & \multicolumn{1}{c|}{0.5} & 0.53 & \multicolumn{1}{c|}{0.46} \\ \hline
 &  &  &  &  &  &  &  &  &  &  &  &  &  &  &  \\
 & \multicolumn{15}{c}{Deep RestoreNET} \\ \cline{2-16} 
\multicolumn{1}{c|}{} & \multicolumn{3}{c|}{Dungeon} & \multicolumn{3}{c|}{Hong-Shevell} & \multicolumn{3}{c|}{White} & \multicolumn{3}{c|}{Gradient} & \multicolumn{3}{c|}{Chevreul} \\ \cline{2-16} 
\multicolumn{1}{c|}{} & \multicolumn{1}{c|}{In} & Out-L & \multicolumn{1}{c|}{Out-R} & \multicolumn{1}{c|}{In} & Out-L & \multicolumn{1}{c|}{Out-R} & \multicolumn{1}{c|}{In} & Out-L & \multicolumn{1}{c|}{Out-R} & \multicolumn{1}{c|}{In} & Out-L & \multicolumn{1}{c|}{Out-R} & \multicolumn{1}{c|}{In} & Out-L & \multicolumn{1}{c|}{Out-R} \\ \hline
\multicolumn{1}{|c|}{R} & \multicolumn{1}{c|}{0.58} & 0.74 & \multicolumn{1}{c|}{0.28} & \multicolumn{1}{c|}{0.58} & 0.31 & \multicolumn{1}{c|}{0.58} & \multicolumn{1}{c|}{1} & 0.93 & \multicolumn{1}{c|}{0.97} & \multicolumn{1}{c|}{0.5} & 0.48 & \multicolumn{1}{c|}{0.55} & \multicolumn{1}{c|}{1} & 0.93 & \multicolumn{1}{c|}{0.91} \\
\multicolumn{1}{|c|}{G} & \multicolumn{1}{c|}{1} & 0.39 & \multicolumn{1}{c|}{0.7} & \multicolumn{1}{c|}{1} & 0.5 & \multicolumn{1}{c|}{0.42} & \multicolumn{1}{c|}{0.5} & 0.51 & \multicolumn{1}{c|}{0.39} & \multicolumn{1}{c|}{0.5} & 0.48 & \multicolumn{1}{c|}{0.46} & \multicolumn{1}{c|}{0.5} & 0.53 & \multicolumn{1}{c|}{0.5} \\
\multicolumn{1}{|c|}{B} & \multicolumn{1}{c|}{0} & 0.49 & \multicolumn{1}{c|}{0.56} & \multicolumn{1}{c|}{0} & 0.49 & \multicolumn{1}{c|}{0.53} & \multicolumn{1}{c|}{0} & 0.036 & \multicolumn{1}{c|}{0.075} & \multicolumn{1}{c|}{0} & 0.074 & \multicolumn{1}{c|}{0.073} & \multicolumn{1}{c|}{0.5} & 0.5 & \multicolumn{1}{c|}{0.48} \\ \hline
\multicolumn{1}{l}{} & \multicolumn{1}{l}{} & \multicolumn{1}{l}{} & \multicolumn{1}{l}{} & \multicolumn{1}{l}{} & \multicolumn{1}{l}{} & \multicolumn{1}{l}{} & \multicolumn{1}{l}{} & \multicolumn{1}{l}{} & \multicolumn{1}{l}{} & \multicolumn{1}{l}{} & \multicolumn{1}{l}{} & \multicolumn{1}{l}{} & \multicolumn{1}{l}{} & \multicolumn{1}{l}{} & \multicolumn{1}{l}{} \\
\multicolumn{1}{l}{} & \multicolumn{15}{c}{\textit{Zhang et al.}} \\ \cline{2-16} 
\multicolumn{1}{c|}{} & \multicolumn{3}{c|}{Dungeon} & \multicolumn{3}{c|}{Hong-Shevell} & \multicolumn{3}{c|}{White} & \multicolumn{3}{c|}{Gradient} & \multicolumn{3}{c|}{Chevreul} \\ \cline{2-16} 
\multicolumn{1}{c|}{} & \multicolumn{1}{c|}{In} & Out-L & \multicolumn{1}{c|}{Out-R} & \multicolumn{1}{c|}{In} & Out-L & \multicolumn{1}{c|}{Out-R} & \multicolumn{1}{c|}{In} & Out-L & \multicolumn{1}{c|}{Out-R} & \multicolumn{1}{c|}{In} & Out-L & \multicolumn{1}{c|}{Out-R} & \multicolumn{1}{c|}{In} & Out-L & \multicolumn{1}{c|}{Out-R} \\ \hline
\multicolumn{1}{|c|}{R} & \multicolumn{1}{c|}{0.58} & 0.59 & \multicolumn{1}{c|}{0.56} & \multicolumn{1}{c|}{0.58} & 0.58 & \multicolumn{1}{c|}{0.59} & \multicolumn{1}{c|}{1} & 1 & \multicolumn{1}{c|}{1} & \multicolumn{1}{c|}{0.5} & 0.52 & \multicolumn{1}{c|}{0.49} & \multicolumn{1}{c|}{1} & 1 & \multicolumn{1}{c|}{1} \\
\multicolumn{1}{|c|}{G} & \multicolumn{1}{c|}{1} & 0.98 & \multicolumn{1}{c|}{0.98} & \multicolumn{1}{c|}{1} & 0.99 & \multicolumn{1}{c|}{0.98} & \multicolumn{1}{c|}{0.5} & 0.5 & \multicolumn{1}{c|}{0.49} & \multicolumn{1}{c|}{0.5} & 0.49 & \multicolumn{1}{c|}{0.51} & \multicolumn{1}{c|}{0.5} & 0.48 & \multicolumn{1}{c|}{0.51} \\
\multicolumn{1}{|c|}{B} & \multicolumn{1}{c|}{0} & 0.027 & \multicolumn{1}{c|}{0.027} & \multicolumn{1}{c|}{0} & 0.02 & \multicolumn{1}{c|}{0.02} & \multicolumn{1}{c|}{0} & 0.012 & \multicolumn{1}{c|}{0.012} & \multicolumn{1}{c|}{0} & 0.012 & \multicolumn{1}{c|}{0.012} & \multicolumn{1}{c|}{0.5} & 0.48 & \multicolumn{1}{c|}{0.51} \\ \cline{1-16} 
\end{tabular}\caption{Input and model responses for the deep CNNs for the different visual illusions studied.}\label{Tableforcolorillusions_deep}
\end{table*}

\subsubsection{Chromatic case}
 Also for chromatic stimuli we perceive physically identical targets as being different, depending on the surround.
 In particular, in the Dungeon stimuli of Fig.~\ref{fig_stimuli1} (third row), the hue of the left target departs from green (to orange), while the target at the right is much greener. 
 In the Hong-Shevell case the target at the right departs away from green. In the case of the White illusion we perceive the left target to be lighter and yellower than the right one. 
 In the color gradient, the targets at the right are seen greener in contrast with the locally red background.
 Finally, in the Chevreul case, each band is perceived to be lighter and darker in its left and right sides respectively.
 
 Figure \ref{fig:colorDNNET} and Table \ref{Tableforcolorillusions}
 show the responses' outputs for the shallow CNNs. In this case, it is more difficult to study them than in the achromatic case, and for this reason we are going to detail the results for each Visual Illusion individually.
 
 \paragraph{\textbf{Dungeon}} In this case, we can see that for the three cases (DN-NET, DB-NET, and RestoreNET), the response for the R channel is larger for the left target (departs from green) and the response for the G channel is much larger in the right target, therefore qualitative matching human perception (the one at the right is \emph{greener}).
 
  \paragraph{\textbf{Hong-Shevell}} For this illusion, we can see that for the three cases (DN-NET, DB-NET, and RestoreNET), the response for the R channel is larger for the right target (departs from green) and the response for the G channel is much larger in the left target, \emph{qualitative matching human perception} (the one at the left is \emph{greener}).
  
    \paragraph{\textbf{White}} For this illusion, we can see that for the three cases (DN-NET, DB-NET, and RestoreNET), the response for the R channel is slighly smaller for the left target and the response for the G channel is much larger in the left target. Therefore, the left target moves towards yellow, while the target at the right is more reddish. Moreover, note that the differences in the G channel are bigger than the differences in the R channel implying not only more yellowish hue, but also brighter target at the left, \emph{qualitative matching human perception} (the one at the left is \emph{brighter and more yellow} vs \emph{darker and more reddish} at the right). 
    
    \paragraph{\textbf{Gradient}} For this illusion, in the three cases (DN-NET, DB-NET, and RestoreNET), the response for the R channel is larger for the right target and the response for the G channel is much larger in the left target. This means that the left target is greener. This implies that there is illusion, but \emph{the illusion goes in the opposite direction of human perception}.
  
    \paragraph{\textbf{Chevreul}} For this illusion, we can see that the effect (the left side of the regions having larger responses than right side) appears in all three color channels for DB-NET, and RestoreNET, but it only appears in the R channel (the brighter one) for the case of DN-Net. Therefore, shallow networks \emph{qualitatively match human perception} also in this case. 
\\
\\
Moving to the deep CNNs studied (including Zhang), Table \ref{Tableforcolorillusions_deep} shows the responses' outputs for them. As with the shallow case we are going to detail the results for each visual illusion individually.
 \paragraph{\textbf{Dungeon (Deep)}} We can see that for Deep DN-NET, Deep DB-NET, and Deep RestoreNET, the response for the R channel is much larger for the left target and the response for the G channel is much larger in the right target, therefore \emph{qualitative matching human perception}. In the case of Zhang, the response for the R channel is slightly larger for the left target, which coincides with human perception, but the effect is very small.
 
  \paragraph{\textbf{Hong-Shevell (Deep)}} We can see that for Deep DN-NET, Deep DB-NET, and Deep RestoreNET, the response for the R channel is larger for the right target and the response for the G channel is larger in the left target, \emph{qualitative matching human perception}. This said, the differences are smaller than in the shallow case, and the effect is therefore attenuated. In the case of Zhang, again, the effect does not appear as all the channels have ``almost'' the same value for both targets.
  
    \paragraph{\textbf{White (Deep)}} In this case, the results for Deep DN-NET, Deep DB-NET, and Deep RestoreNET follow the same trend than their shallow counterparts - R channel slightly smaller for left target, G channel larger for left target- but with smaller differences, meaning than the effect -the left side perceived yellower- is still present but at a reduced rate. Again in this case, the CNN from Zhang does not produce any effect as it does not almost modify the input values.
   
    \paragraph{\textbf{Gradient (Deep)}} In this case, Deep DN-Net, Deep DB-NET and Deep RestoreNET output a higher value for the G channel in the left target, and, although deep DN-NET outputs a higher value in the R channel for the left target, this is probably not enough to mimic the human perception effect. In contrast, in this case, Zhang outputs shift in the same direction as humans. 
  
      \paragraph{\textbf{Chevreul (Deep)}} In this particular case, the results for the Deep DN-NET, Deep DB-NET and Deep RestoreNET are very similar to their shallow counterparts, meaning that they roughly have a brightness shift in the human-like direction. Finally, again, the CNN from Zhang does not present the effect, because the value of the right target in the G and B channels is higher than the left one.\\

To summarize, we have seen that the simple CNNs for basic image processing tasks that we have studied in this paper qualitatively match the human perception in around $80\%$ of the cases. This striking rate led us to study them in a more standard vision science setup, such as the correspondent pairs paradigm in which we focus on the next subsection. Also, it is remarkable that a very recent CNN tailored at performing state-of-the-art denoising does not modify the input images much, it is able to reconstruct the images almost perfectly and therefore this clearly attenuates or makes disappear the visual illusion effects.

\subsection{Corresponding pairs in color induction}

\begin{figure*}
\begin{center}
\hspace{-0.5cm}\includegraphics[width=0.9\linewidth]{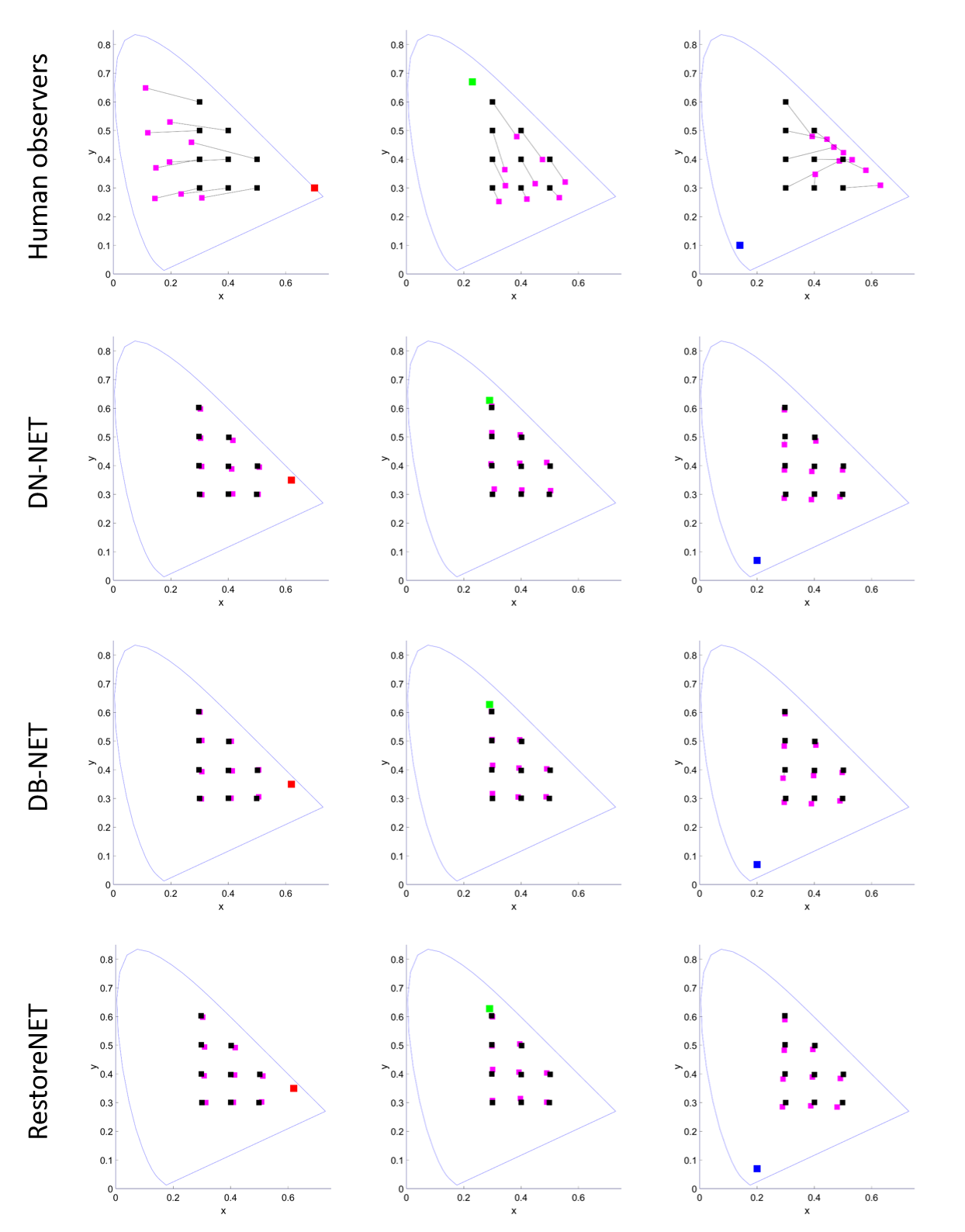}
\end{center}
\captionof{figure}{First row: Results for the human observers in the Ware-Cowan corresponding pair experiment. Second to last raw: Results for the shallow CNNs studied in the Ware-Cowan corresponding pair experiment. We can see that the displacements are small, and in the opposite direction than for human observers (they suffer from \emph{assimilation} with the inductor as opposed to \emph{contrast} happening in human observers). Note that the inductors used by Ware \& Cowan in their psychophysical experiments are slighly more saturated than those used in our \emph{numerical psychophysics}. This is because we were using images expressed in digital counts. Nevertheless, this small difference in the inductors does not justify the differences in the corresponding pairs. Therefore, qualitative conclusions about the differences of behavior between networks and humans are valid.}
\label{results_wc_shallows}
\end{figure*}

\begin{figure*}
\begin{center}
\hspace{-0.5cm}\includegraphics[width=0.95\linewidth]{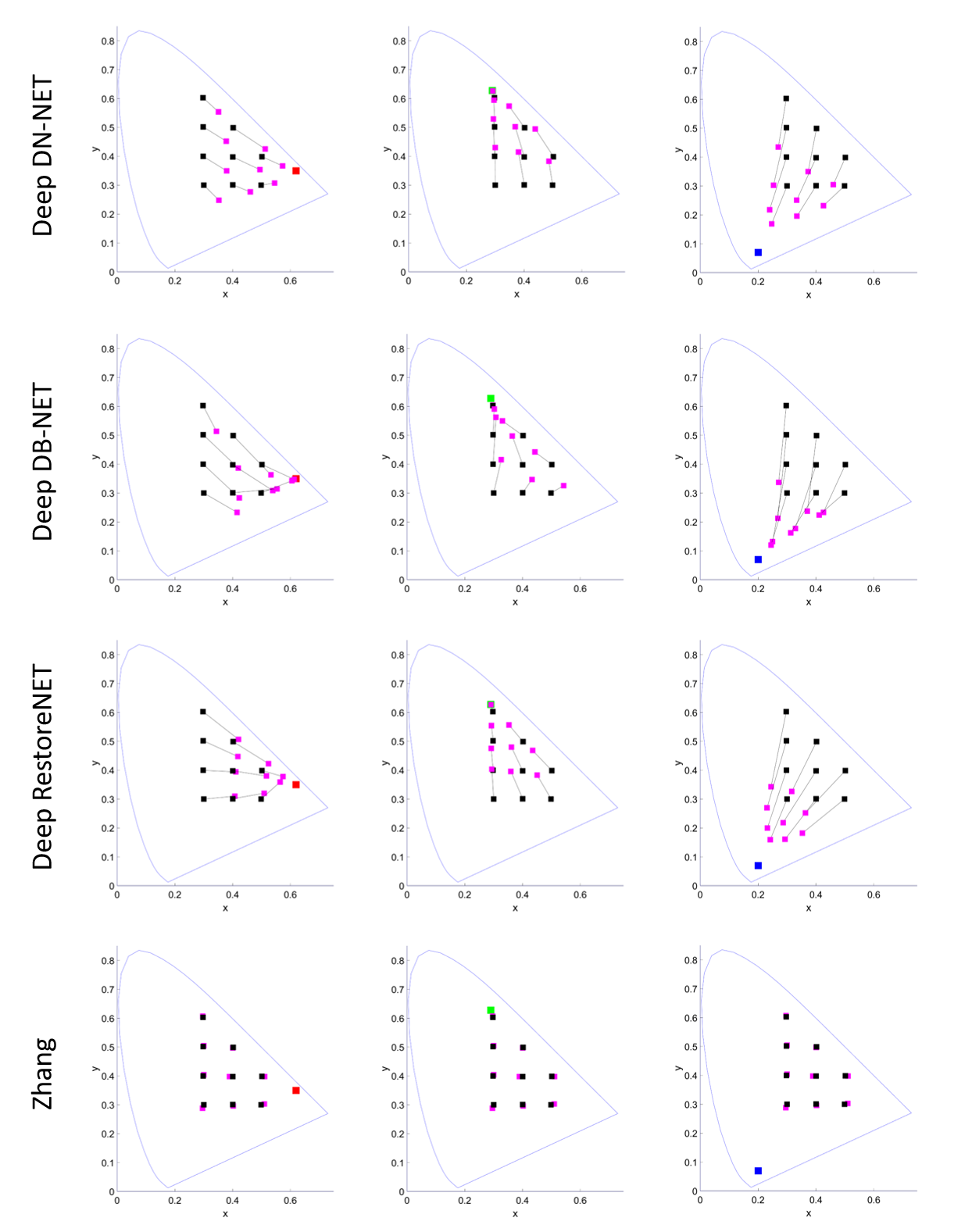}
\end{center}
\captionof{figure}{Results for the deep CNNs studied in the Ware-Cowan corresponding pair experiment. We can see that the displacements of the 4 layer networks (Deep DN-NET, deep DB-NET, and deep RestoreNET) are comparable in magnitude to the human illusions, but again in the opposite direction than for human observers. 
The (really deep) network of Zhang has the smallest displacements of all explored networks. However, in some cases we got contrast (departure from inductor) as in human observers, but in any case, illusions of very small magnitude.}
\label{results_wc_deeps}
\end{figure*}

In this experiment, given a fixed test-surround configuration, $\mathbf{i} = [\mathbf{t} \,\,\,\, \mathbf{s}]^\top$, the \emph{observer} looks for the \emph{corresponding pair}, $\mathbf{t'}$, that seen on a neutral reference background, $\mathbf{w}$, matches the perception of $\mathbf{t}$. 

While human observers look for the corresponding pair by  physically changing the color in the lab, we say that the network matches the perception when, given these two responses, 
\begin{equation}
\left[\!\!
\begin{array}{c}
\mathbf{r}_t\\
\mathbf{r}_s\\
\end{array}
\!\!\!\right] = m\left( \,\left[\!\!
\begin{array}{c}
\mathbf{t}\\
\mathbf{s}\\
\end{array}\!\!\right] , \theta_k \right)
\,\,\,\,\,\,\textrm{and}\,\,\,\,\,\,
\left[\!\!
\begin{array}{c}
\mathbf{r}_{t\mathbf{'}}\\
\mathbf{r}_w\\
\end{array}\!\!\!
\right] = m\left( \,\left[\!\!
\begin{array}{c}
\mathbf{t'}\\
\mathbf{w}\\
\end{array}\!\!\right] , \theta_k \right)
\end{equation}
it holds $\mathbf{r}_t = \mathbf{r}_{t\mathbf{'}}$.
Therefore, the cost function for the \emph{numerical} corresponding pair experiment is just:
\begin{equation}
     \mathbf{t'} = \argminA_{t^\star} \,\,|\,\, \mathbf{r}_{t^\star}( [\mathbf{t^\star} \,\,\, \mathbf{w}] ) - \mathbf{r}_{t}( [\mathbf{t} \,\,\, \mathbf{s}] ) \,\,|_2
\end{equation}

Results for this experiment are shown in Figures \ref{results_wc_shallows} and \ref{results_wc_deeps}. In these figures, the black squares represent the test colours, while the magenta squares represent the corresponding pair required by the model. The red, green and blue squares are the inducers. The first row in Figure \ref{results_wc_shallows} presents the results obtained by the human observers \cite{ware82}. The rest of rows in Figure \ref{results_wc_shallows} show the results for the shallow CNNs, while Figure \ref{results_wc_deeps} presents the results for the deep CNNs. In these Figures we can clearly see that the different CNNs produce some effect in the starting value -specially in the case of the Deep CNNs where the displacements are of the same magnitude than that of humans-. However, the effect that these nets produce is actually the opposite of the one measured for human observers, with the except of the CNN from Zhang, in which some of the patches -in particular those that are further from the inducer- follow the color contrast trend found in human observers. Interestingly, this result relates to the results obtained for the Gradient illusion in the previous subsection, where the CNN from Zhang was the only one presenting the effect (if mildly). This connection was to be expected, as the Gradient illusion is
the closest one to the Ware-Cowan experiment
of all the illusions studied there.

Results of this corresponding pair experiment
 can be summarized as follows. 
(1)~Simpler networks do have illusions (particularly the 4-hidden layer architecture), and the really deep (17-layer) network of Zhang presents the smallest amount of illusion. 
(2)~However, the illusions of the simpler networks go in the assimilation-by-inducer direction, as opposed to the contrast effect seen in humans.  

\section{Linear analysis of the results}

Results of the numerical experiments with the networks trained for visual tasks confirm that they do have illusions, as anticipated  before~\cite{GomezCVPR19}, but they do not necessarily have the same illusions as the humans. This section analyzes why, particularly for the assimilation effect seen in Ware-Cowan experiments.

Here we show how the behavior of visual CNNs can be understood through the analysis of the linear approximation of their response.
Specifically, consider the following first order approximation of Eq.~\ref{response}:
\begin{equation}
     \mathbf{r} = m(\mathbf{0},\theta_k) + \nabla_{\mathbf{i}}\, m(\mathbf{0},\theta_k) \cdot \mathbf{i} = M_{\theta_k} \cdot \mathbf{i}
     \label{eq:linear_approx1}
\end{equation}
where the matrix $M_{\theta_k}$ is the Jacobian of the network response w.r.t. the input at $\mathbf{0}$, and we assume that the response for the null stimulus is also zero. This matrix is fixed for a certain architecture/task combination.

The Jacobian for different points gives important information about the behavior of the network and could be computed analytically for any input~\cite{Martinez18}. However, for our purposes here (we need it only at the origin, $\mathbf{0}$), it can be obtained through plain linear regression. Once some architecture has been trained for certain task, the resulting model, characterized by the parameters $\theta_k$, can be applied to a set of $N$ stimuli. Then, by stacking the vectors representing the $N$ stimuli and the $N$ responses in the matrices, $\mathbf{I} = [\mathbf{i}^{(1)} \, \mathbf{i}^{(2)} \cdots \mathbf{i}^{(N)}]$, and  $\mathbf{R}_{\theta_k} = [\mathbf{r}^{(1)} \, \mathbf{r}^{(2)} \cdots \mathbf{r}^{(N)}]$, respectively, we have:
\begin{equation}
    M_{\theta_k} = \mathbf{R}_{\theta_k} \cdot \mathbf{I}^\dag
    \label{eq:linear_approx2}
\end{equation}
where $\mathbf{I}^\dag$ is the pseudoinverse of the rectangular matrix with the input images.

While CNNs are, in general, difficult to understand~\cite{samek2019explainable}, if the proposed approximation captures a substantial fraction of the energy of the response, it can be very useful for two reasons: (1) it allows the use of well understood linear algebra in the analysis, and (2) it allows the comparison with classical linear descriptions of human vision.

In particular, here we perform two kinds of linear analysis, where the second is justified by the results of the first:
\begin{enumerate}
    \item  First, we make no extra assumptions (apart from linearity) and we perform an eigen-vector analysis of the matrix $M_{\theta_k}$.
    This analysis shows that this kind of networks
    are stationary (shift invariant), they are roughly spatio-chromatically separable, they implicitly operate in chromatic opponent spaces, and they have markedly different spatial bandwidth in these chromatic channels.
    \item The above results imply that extra assumptions can be done on top of linearity and hence, they justify an analysis of the transfer functions of the networks in the Fourier domain of conventional opponent channels.
\end{enumerate}

Before doing the above linear analysis, we qualitatively illustrate the accuracy and the basic spatial property of the linear approximation. Then we address points 1 and 2. Finally, we come back to the quantification of the nonlinear nature of the networks in discussing the amount of illusion depending on their deepness (Table~\ref{tableta}).

\begin{figure*}
\begin{center}
\includegraphics[width=1.0\linewidth]{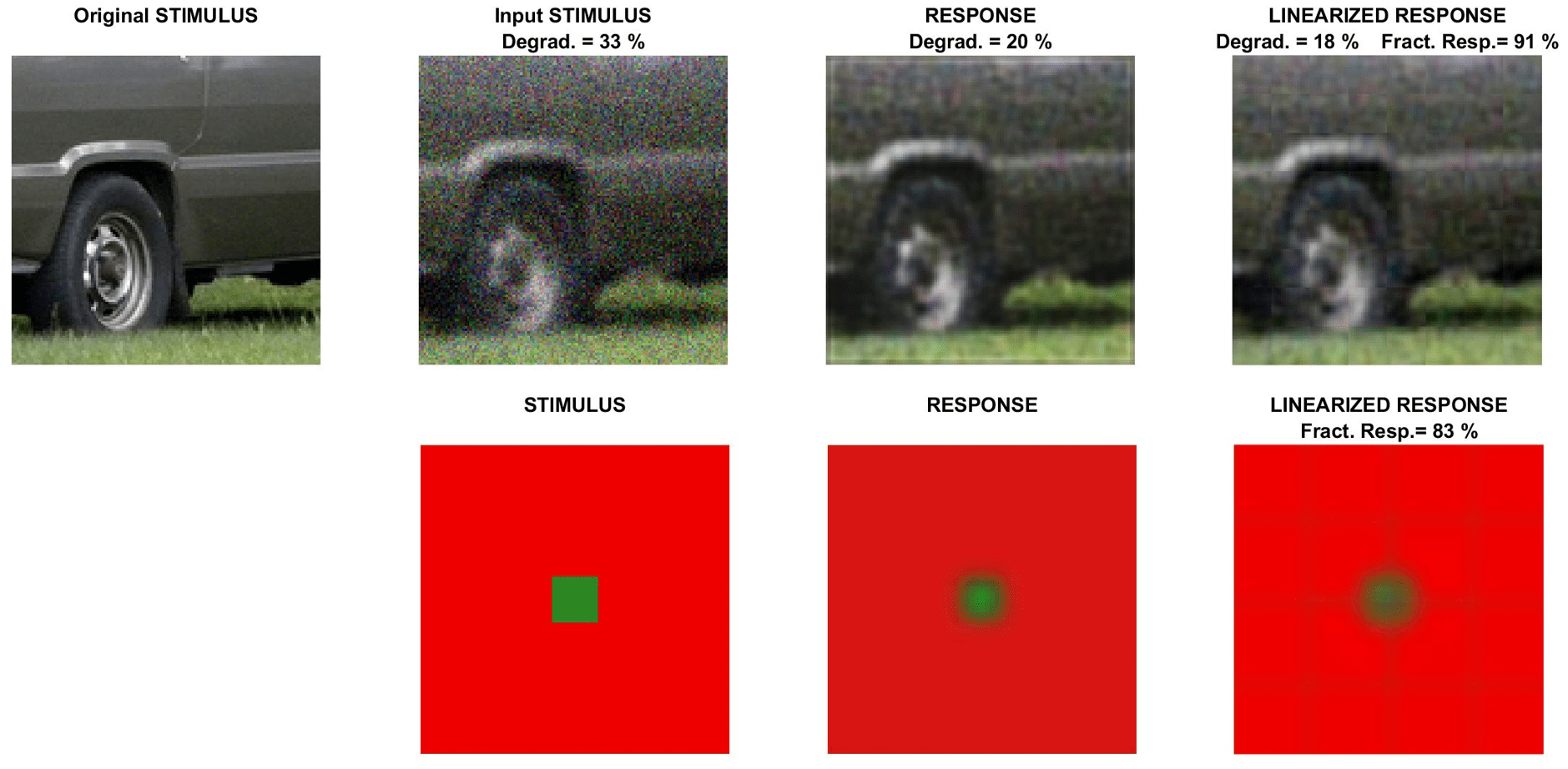}
\end{center}
\vspace{-0.4cm}
\caption{Representative response of RestoreNET on a natural image and on a illusion inducing stimulus. In the restoration example (top) \emph{Degradation} represents the RMSE of the considered signal (either input signal or response -restored signal-) referred to the square root of the average energy of the original stimulus.
The \emph{Fraction of Response} is the RMSE difference between the linearized response and the actual (nonlinear) response referred to the square root of the average energy of the network response.}
\label{is_linear}
\end{figure*}

\subsection{Linear approximation is representative and shift invariant}

We applied Eq.~\ref{eq:linear_approx2} to $1.3 \cdot 10^5$ image patches subtending 0.23 deg ($16\times16$ pixels, i.e. vectors $\mathbf{i}$ and $\mathbf{r}$ of dimension $16\times16\times3 = 768$). Therefore this specific illustration of Eq.~\ref{eq:linear_approx2} required the pseudoinverse of a matrix of size $768\times1.3\cdot 10^5$.
Figure \ref{is_linear} shows a representative example that illustrates the accuracy of the linear approximation. We show the behavior of the shallow RestoreNET with the kind of images used in the training and with the kind of stimuli used in the simulations of the visual illusions.

\begin{figure*}
\begin{center}
\includegraphics[width=1.1\linewidth]{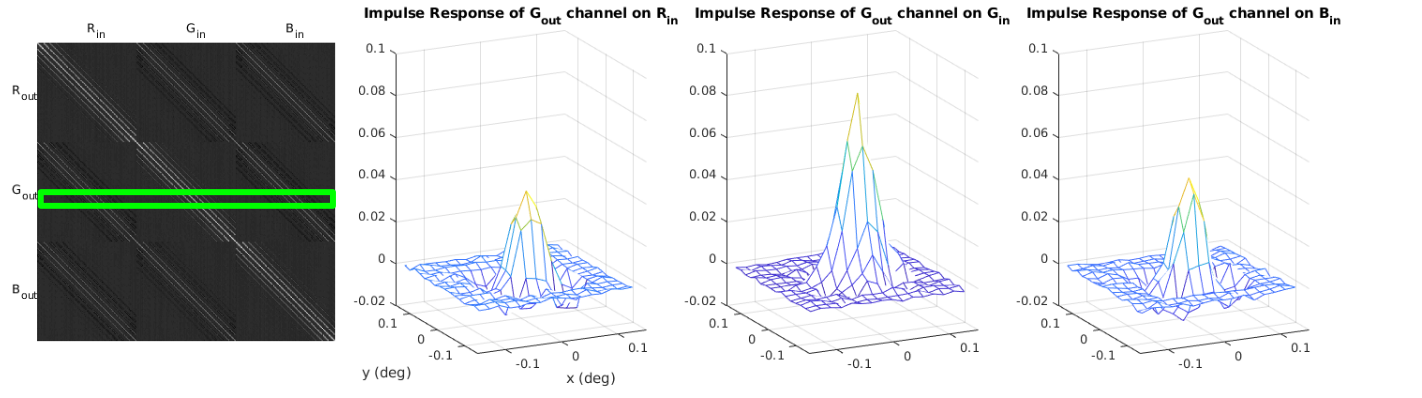}
\end{center}
\vspace{-0.4cm}
\caption{Linearized version of RestoreNet: the matrix $M_{\theta_k}$.
In order to interpret the matrix (at the left), remember that each RGB channel of the input color images was spatially vectorized, and vertically stacked one after the other. Therefore, in the linear approximation of Eq.~\ref{eq:linear_approx1}, each row of this matrix acts on the image column vector that was arranged in this specific way.
Large submatrices represent the spatial processing within each color channel and then, these responses are linearly combined to lead to the final responses.
The specific highlighted row corresponds to an output unit that is tuned to a the central location of the image patch and it is mainly tuned to (it receives more input from) the G channel. This can be better seen by spatially rearranging the weights to be applied to the R, G, and B inputs, which is the result displayed at the surfaces.}
\label{LinearAprox}
\end{figure*}

\begin{figure*}[t!]
\begin{center}
\includegraphics[width=0.9\linewidth]{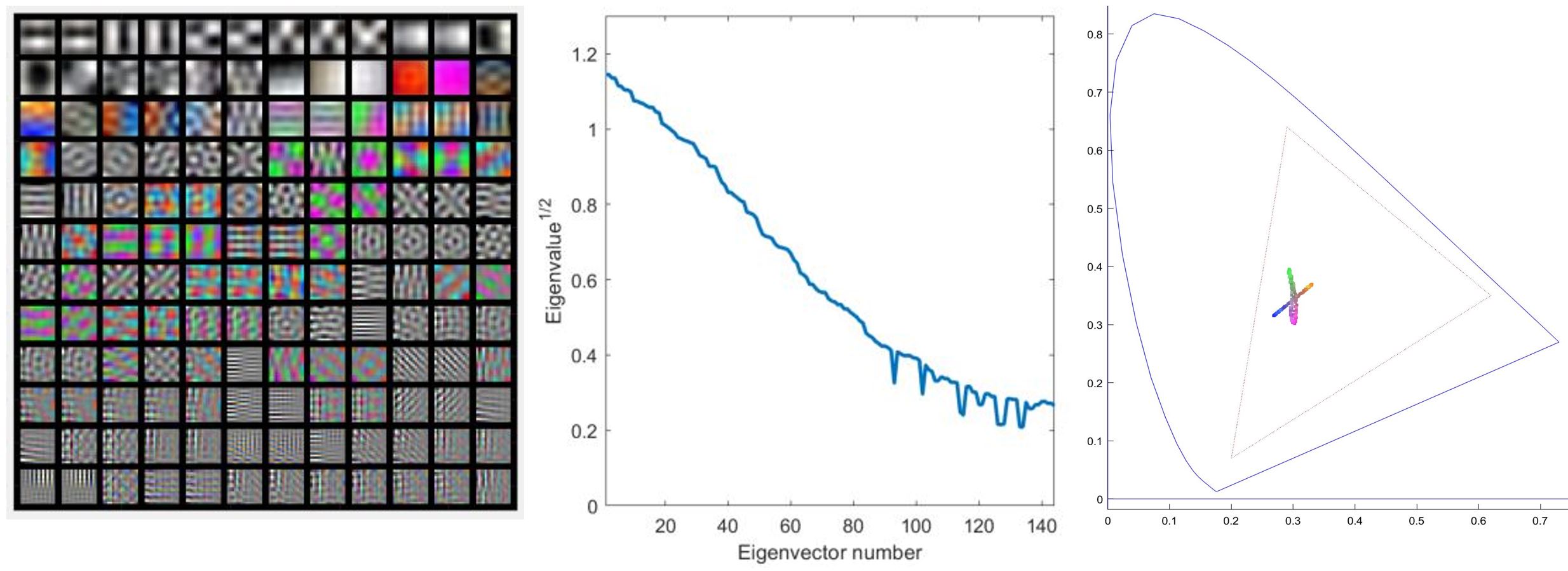}
\end{center}
\caption{Eigenfunctions (left) and eigenvalues (center) of the linearized network. Eigenfunctions (columns of $B$) are sorted sorted (left-to-right and top-to-bottom) according to the associated eigenvalues. The CIE xy chromatic diagram (right) displays the chromatic coordinates of the colors in the 50 eigenfunctions with bigger eigenvalue.
Note that as the input stimuli in the considered dataset are given in digital counts the eigen-functions are also given in digital counts. Therefore, the representation of the colors of the eigen-funcitons in the chromatic diagram involved the assumption of a standard display calibration conversion~\cite{Colorlab}.
Note also that, given the fact that $M_{\theta_k}$ is almost symmetric (see Fig.~\ref{LinearAprox}), $B$ is almost orthonormal and hence, the \emph{encoding} functions (rows of $B^{-1}$, not shown) are very similar to the \emph{decoding} functions represented here.}
\label{eigen1}
\end{figure*}

In this example we see that:
(1) the network carries out the visual task (i.e. it is reducing the degradation),
(2) the network visually behaves as classical restoration techniques (e.g. Wiener, Tikhonov~\cite{wiener1949extrapolation,tikhonov1977solutions}, see \cite{Gutierrez06} for visual examples).
(3) The network seems to have a stationary behavior, which may not be surprising given the stationary nature of the degradation learnt.
(4) Given the above, a linear version of the network may be sensible.
(5) The intuitive meaningfulness of the linear approximation is confirmed by the results shown at the right column.
First, note the visual resemblance of the actual and linear responses, particularly in natural images (where $M_{\theta_k}$ came from). Second, note the large fraction of energy of the response captured by the linear approximation in these particular images (for a larger dataset of natural images the figure is about 94\%). Incidentally, for this specific image the linear approximation gives a slightly better restoration result, but this is not representative of the whole natural image dataset.

Fig.~\ref{LinearAprox} explicitly shows the matrix $M_{\theta_k}$ for the RestoreNET example considered in Fig.~\ref{is_linear}.

This result shows
(1)~the existence of well defined submatrices in $M_{\theta_k}$ that correspond to similar spatial processing in the different chromatic channels. This suggests that the behavior of the network maybe \emph{roughly} separable in chromatic and spatial terms.
(2)~the Toeplitz-like structure of the submatrices, that confirms the spatially stationary (roughly convolutional-like) behavior of the network.
(3) the \emph{equivalent} convolution kernels (\emph{equivalent} receptive fields) of the network are a combination of a Gaussian-like blurring operator for the channel at hand, and center-surround operators at adjacent channels.

These features qualitatively resemble the properties of LGN cells, but additional insight is definitely required.
Diagonalization of the matrix $M_{\theta_k}$ done in next section
helps to obtain extra intuition on the inner working of the network.

\subsection{Eigenvector / eigenvalue analysis}

The eigen-decomposition of the linear transform $M_{\theta_k}$ identifies the stimuli that are considered by the system in a \emph{special} way,
By definition, the eigen-functions, $\mathbf{b}^{(i)}$, are stimuli whose response is just an attenuated version of the input: $\lambda_i \mathbf{b}^{(i)} = M_{\theta_k} \cdot \mathbf{b}^{(i)}$, and hence, $M_{\theta_k} = B \cdot \lambda \cdot B^{-1}$, where $B = (\mathbf{b}^{(1)} \, \mathbf{b}^{(2)} \cdots \mathbf{b}^{(d)})$.
Moreover, the eigen-decomposition ranks
the eigen-functions according to the eigen-values.

Therefore, the eigendecomposition describes the response of the network as a \emph{linear autoencoder}:
\begin{equation}
\mathbf{r} = B \cdot \lambda \cdot B^{-1} \cdot \mathbf{i}
\end{equation}
where the \emph{rows} of $B^{-1}$ contain the \emph{encoding functions}, and the \emph{columns} of $B$ contain the \emph{decoding functions}.

In this interpretation of the action of the network, the \emph{encoder}, $B^{-1}$, transforms the input stimuli into a new representation. This is the inner eigen-representation of the network.
In this inner representation, coefficients of the signal are dimension-wise attenuated by the diagonal matrix $\lambda$, and then, the final response is synthsized by the \emph{decoder} $B$.

Fig.~\ref{eigen1} shows the eigen-functions (columns of $B$) of the considered $M_{\theta_k}$. The most relevant stimuli for the network appear first.

The diagonalization of $M$ shows that:
(1)~Eigenfunctions are oscillating stimuli of different frequencies extended over the spatial domain (stationary textures over the spatial domain).
(2)~Oscillations appear on the achromatic direction and in two \emph{very specific} chromatic directions: namely \emph{pink/green}, and \emph{yellow-orange/blue}.
(3)~The most important functions are the achromatic and only afterwards there are functions that display chromatic variations (but also brightness oscillations of different frequency).
These facts strongly suggest that the network is \emph{implicitly} analyzing the stimuli in a Fourier-like (or DCT-like) representation in
a color opponent space.

In order to clarify this intuition, we did the following analysis:
first we computed the change of basis matrix that transforms the CIE XYZ primaries into the color basis defined by the extreme
colors of the pink/green, yellow-orange/blue, and dark/light gray directions found in the eigenfunctions.
This matrix allows to compute the color matching functions in the new basis.
The perceptual meaningfulness of the \emph{intrinsic} color basis of the network is demonstrated in Fig.~\ref{opponent}.

\begin{figure*}
\begin{center}
\includegraphics[width=1.0\linewidth]{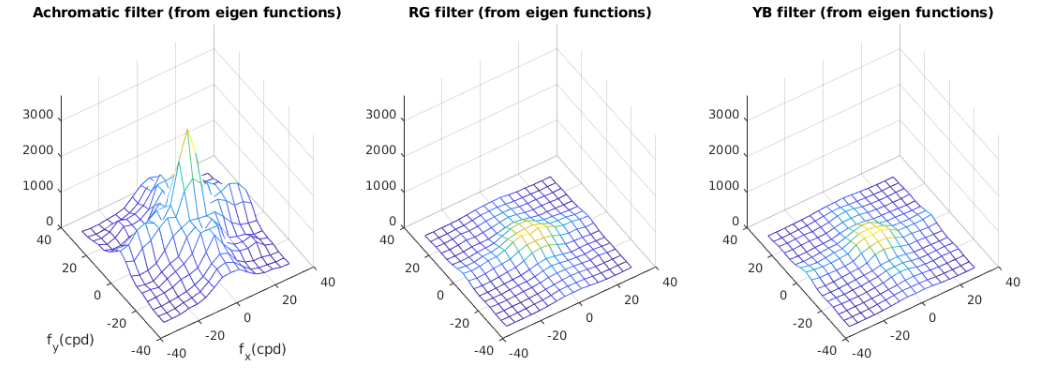}
\end{center}
\caption{Accumulated spectra of eigenfunctions decomposed in their intrinsic color space and weighted by eigenvalues. Limited frequency resolution is due to the fact that this result comes from small $16\times16$ image blocks. This may give rise to artifacts in the spectra.}
\label{eigen2}
\end{figure*}

Then, in order to estimate the spatial bandwidth of the network in these chromatic channels just found,
we accumulated the spectra of the eigenfunctions decomposed in this color space and weighted the spectra by the corresponding eigenvalues.
This is how images decomposed in this color space would be weighted when passing through the network.
The result of such analysis is shown in Fig.~\ref{eigen2}.
According to that, the intrinsic representation of the network can be interpreted as a color decomposition of stimuli in certain
opponent color space (which is similar to human opponency), and the application of filters of markedly different bandwidth
in the achromatic and the chromatic channels.

\begin{Figure}
\begin{center}
\hspace{-0.5cm}\includegraphics[width=1.1\linewidth]{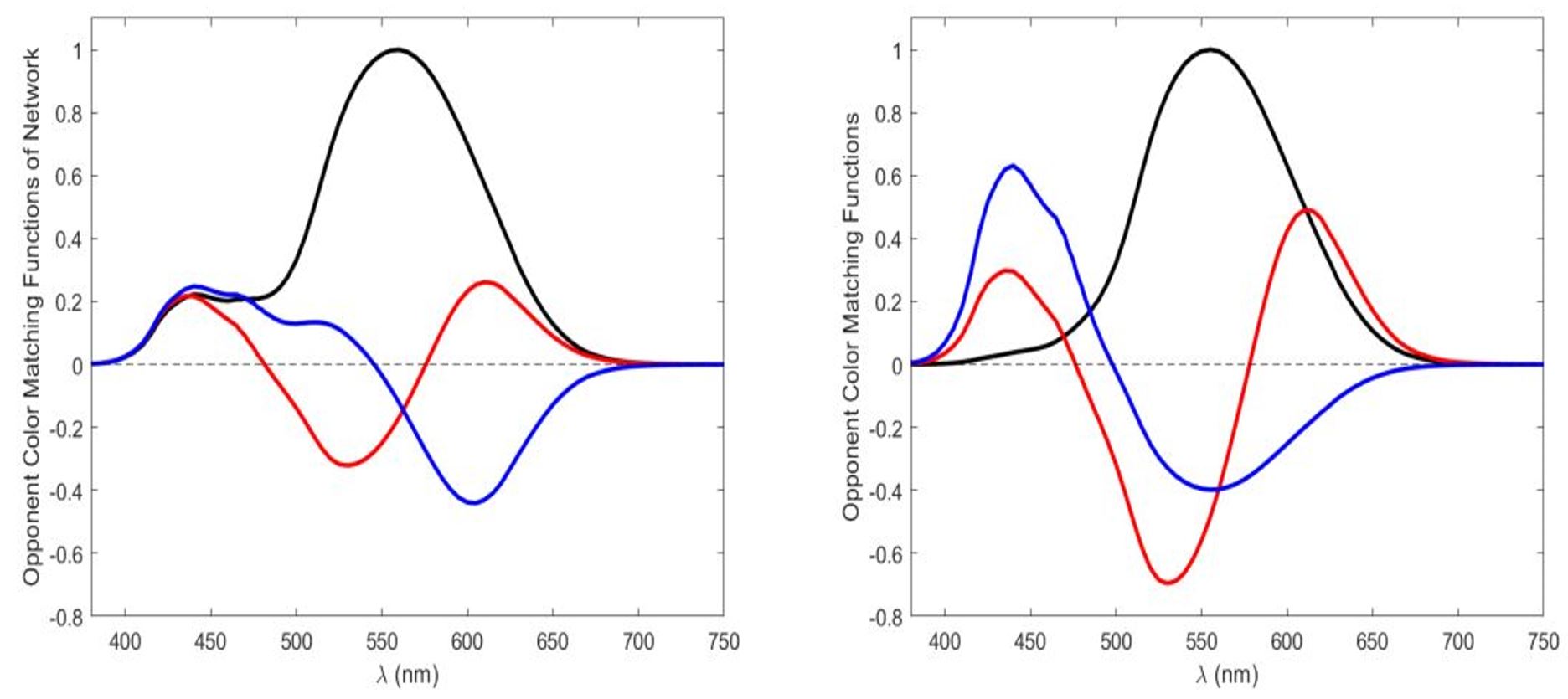}
\end{center}
\captionof{figure}{Intrinsic color matching functions of the network (left) compared with the classical opponent color-matching functions of human color vision (by Jameson and Hurvich~\cite{Jameson59}, on the right). Both systems of primaries have an achromatic channel (all-positive color matching function in black), and two opponent chromatic channels (with positive and negative values).}
\label{opponent}
\end{Figure}

These filters could be compared with the achromatic and chromatic Contrast Sensitivity Functions (CSFs) of human
viewers~\cite{campbell1968application,mullen1985contrast},
but the frequency resolution of this eigen-analysis is limited by the size of the image patches
used in computing the matrix $M_{\theta_k}$.

Note that the only assumption or approximation made so far is \emph{linearity}.
Fortunately, the properties of $M_{\theta_k}$ and $B$ found in the network allow us to make
extra assumptions beyond linearity that make possible a more accurate analysis.

\subsection{Spatial Fourier analysis in opponent channels}

\begin{figure*}
\begin{center}
\includegraphics[width=1\linewidth]{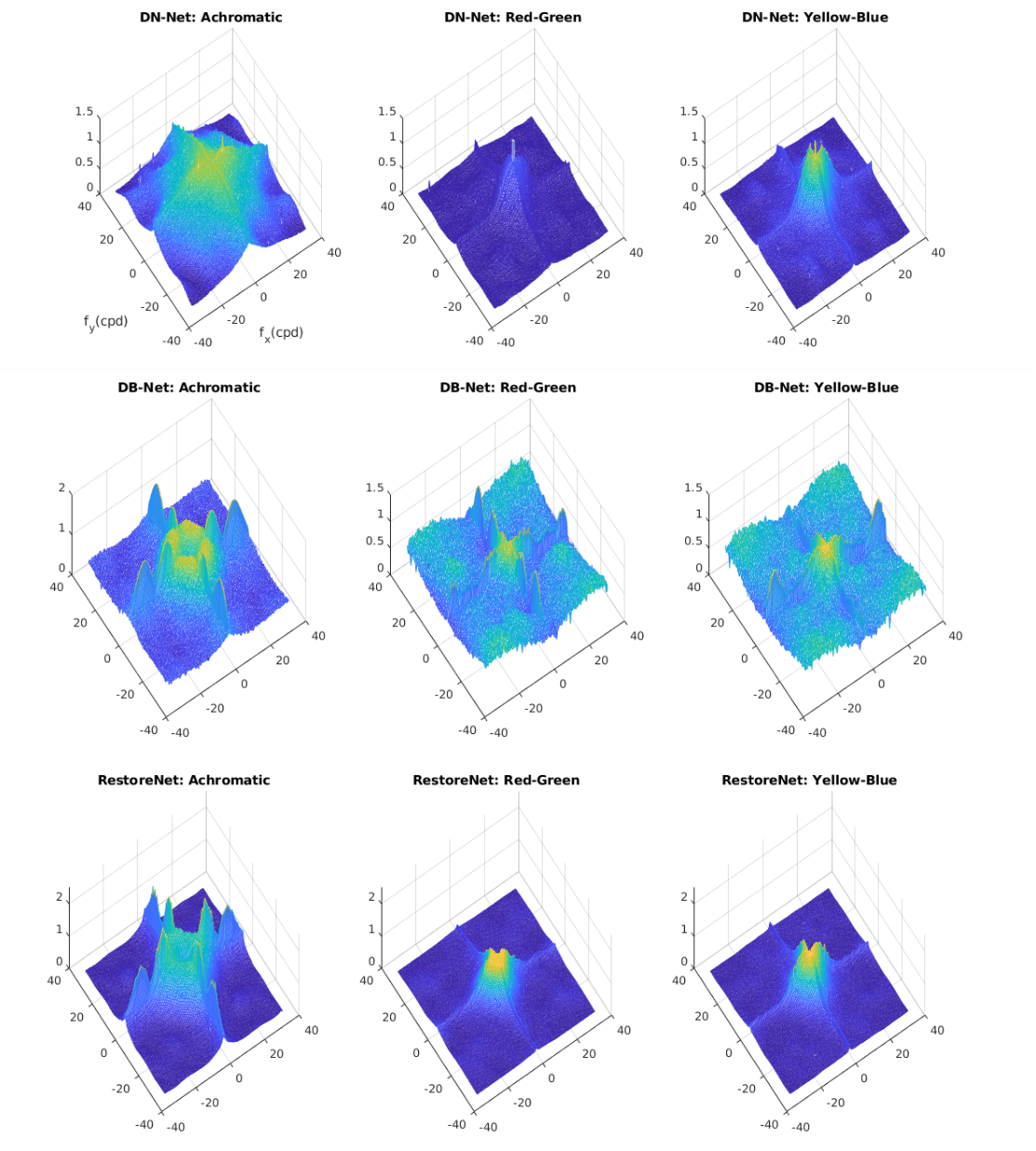}

\end{center}
\caption{Filters estimated for DB-net, DN-net and RestoreNet assuming a classical opponent space and assuming the Fourier representation.
The filters for the deeper version of these networks are similar.
Note that, following classical ideas from optimal filtering and regularization~\cite{wiener1949extrapolation,tikhonov1977solutions}, the denoising filters happen to be low-pass, the deblurring filters happen to be highpass, but of course, also preserving the low-frequencies, and the restoration filters are a combination of both.}
\label{Fourier1}
\end{figure*}

\begin{figure*}[t]
\begin{center}
\includegraphics[width=1.0\linewidth]{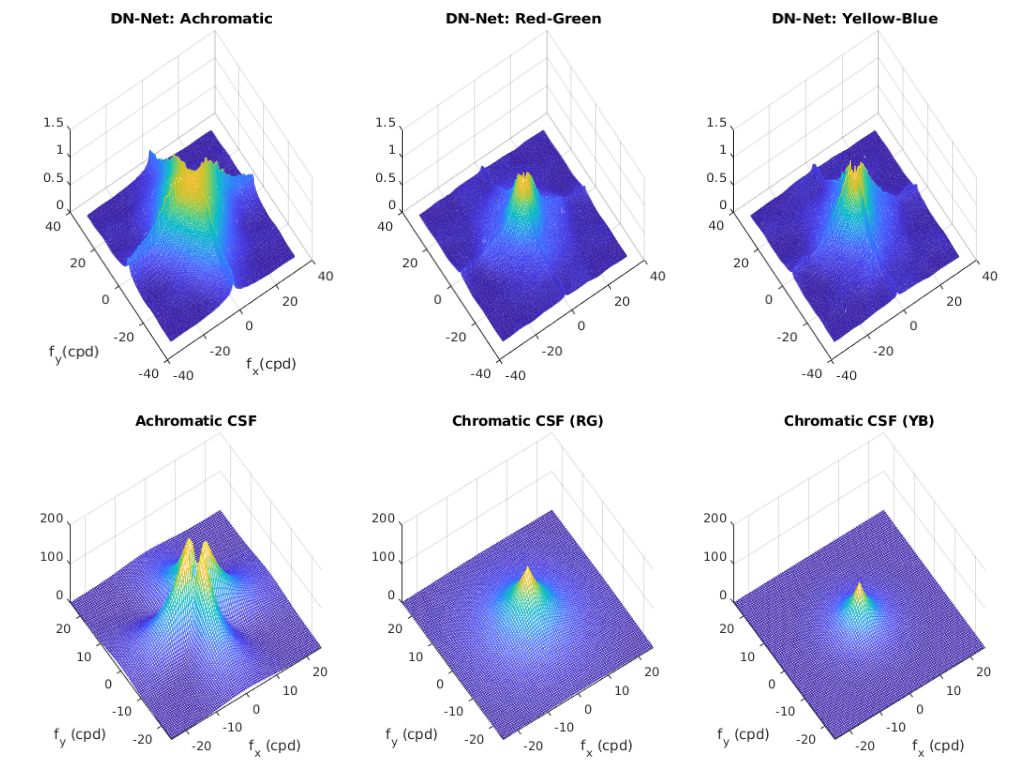}
\end{center}
\caption{Top row: Filters estimated for DB-net, trained with images coming from two calibrated databases (images in CIE XYZ with no spatial manipulation). Results are qualitatively the same that for the large uncalibrated dataset. Bottom row: Human CSFs for convenient reference (Achromatic CSF from the Standard Spatial Observer~\cite{Watson02}, and chromatic CSFs from~\cite{mullen1985contrast}). }
\label{csfs}
\end{figure*}

The properties found above for linear approximations using small-size image patches justify a straightforward
Fourier analysis of the transfer functions of the network.
After finding that the network \emph{implicity} operates in an opponent color space and it is shift-invariant
or stationary, and hence it has eigenfunctions which are Fourier-like, we did the following analysis.
For 2046 full-size images subtending 1.83 deg ($128\times128$ pixels) we computed the quotient of the Fourier spectra
of the input stimuli and the output responses, both decomposed in a classical opponent space~\cite{Jameson59},
the one of the color matching functions represented in Fig.~\ref{opponent}-right.
In this way, the filters will be directly comparable to the human CSFs.
These filters are shown in Fig.~\ref{Fourier1}, and the actual CSFs are plotted for convenient reference in the bottom row of Fig.~\ref{csfs}.

In this context in which models and linear approximations are obtained from large image databases an important safety check was necessary. 
For this specific illustration we not only trained the networks with images from the massive database CLS-LOC~\cite{russakovsky2015imagenet} (uncalibrated images and eventually subject to uncontrolled manipulations), but we also did a separate training with images coming from two calibrated databases (images in CIE XYZ with no spatial manipulation~\cite{Parraga09,Parraga09b,Laparra12,Gutmann14}). Results were qualitatively the same (see Fig.~\ref{csfs} first row). This implies that the database CLS-LOC can be trusted with regard to the average spatio-chromatic spectra (covariance matrix) of the image samples.

\subsection{Linear approximation and strength of illusions}

The linear approximation of the simpler networks (2-hidden layers and 4-hidden layers) reveals a number of human-like characteristics in their intrinsic image representation, namely the chromatic opponent channels and filters of bandwidths similar to the CSFs.

The emergence of this specific frequency selectivity to fulfill the low-level visual task explains that color and luminance profiles in the stimuli are distorted in the response of the network in specific ways.
The responses at certain region changes depending on the spatial context (e.g. Figs.~\ref{fig:grayVI}-\ref{fig:colorDNNET}), thus leading to visual illusions in these networks.

The emergence of these properties can be understood from the relation of image restoration methods with the statistics of natural images: if the goal is removing non-natural features, the network will learn transfer functions matched to the statistics of the signal to filter out undesired features. And it is known that the covariance of natural colors~\cite{Buchsbaum83,sejnowski2002} and natural color images~\cite{Sejnowski2001} is consistent with the opponent color representation and eigenfunctions found by our linear analysis.

As a by-product, the effect of the context due to the width of the equivalent kernels of the linearized network explains the shifts in perceived brightness in similar directions to that happening in humans. But the simplicity of this behavior, e.g. as in Fig.~\ref{is_linear}, 
also explains why the illusions may be markedly different from those of humans in some color cases.

For instance, when the spatial layout of the stimulus is relatively simple, as in the \emph{Gradient color} illusion or in the center-surround setting in the Ware-Cowan experiment, the simple filtering found in Fig.~\ref{is_linear} only leads to penetration of the surround in the region corresponding to the target/test thus leading to \emph{assimilation} instead of the \emph{contrast} found in human observers. 
This eventually simple behavior revealed by the linear analysis certainly applies to the six models with not that many layers (DN-Net, DeepDN-Net, etc.).
However, the situation in much deeper state-of-the-art networks (e.g. the 17 layer architecture of Zhang et al.~\cite{zhang2017beyond}) can be different.

We argue that making the models deeper will (of course) lead to better performance in the specific visual task used in training, and eventually higher nonlinear nature, see Table~\ref{tableta}, but \emph{not necessarily} better resemblance to human viewers.

Note that in Table~\ref{tableta}, as in Fig.~\ref{is_linear}, the \emph{Fraction of Linear Response} describes the nonlinear nature of the network because it represents the fraction of response that can be explained by the linear approximation. The \emph{Performance} (\emph{error} in denoising) is the fraction of clean signal not recovered by network, and the final row qualitatively describes the  magnitude of illusions found.

\bigskip
\begin{minipage}{\linewidth}
\centering
\captionof{table}{Nonlinearity, Performance \& Illusion Strength} \label{tableta} 
\begin{tabularx}{\linewidth}{@{} C{3cm} C{1cm} C{1cm} C{1cm} @{}}\toprule[1.5pt]
                & \bf Shallow & \bf Deep & \bf Zhang et al. \\\midrule
   \bf Fract. Lin. Resp.     &   94\%       & 95\%      &   40\%             \\
   \textbf{Performance} (error)  &  11\%       & 10\%      &    2\%             \\\midrule
   \bf Illusion Strength &   ++         &  +++      &    -             \\   
\bottomrule[1.25pt]
\end {tabularx}\par
\end{minipage}
\bigskip

In the specific case of the simpler networks explored (2-hidden layers and 4-hidden layers respectively), the increased complexity does not make a big difference in terms of their nonlinear nature, which explains the similarity of their intrinsic filters and the slight improvement in performance for the deeper net.
However, the differences are substantial when the flexibility is really increased by using many layers as in Zhang el al.

From a pure machine learning perspective, it is obvious that the nonlinear nature of the networks and their performance in the goal have to increase when substantially increasing the number of parameters. 
However, regarding the magnitude of the illusions (and more in general, regarding the eventual similarity with the visual system) this does not necessarily increase with the complexity of the model.

This can be understood in two ways: (1)~increasing the complexity usually leads to systems that are too specialized in the specific goal. Therefore, it is reasonable that Zhang's network does not have visual illusions with the considered stimuli because these stimuli have no noise. (2)~More interestingly, another way to see the increase of L+NL layers is as actual appropriateness of the considered model in terms of similarity with the visual system. While using a low number of L+NL layers to model this low-level (or early-vision) task may be reasonable from visual physiology point of view, using 17 L+NL layers may be unreasonable. Therefore, it is normal that an inappropriate model leads to non-human behavior (almost no illusion).

\section{Discussion and Final Remarks}

This work confirms and expands our original report on color and brightness illusions suffered by CNNs trained to solve low-level visual tasks~\cite{GomezCVPR19}.
Specifically, we explored a range of five classical brightness illusions and their color counterparts (a total of 10 different illusions) to point out the existence of illusions in CNNs and assess their qualitative correspondence with human behavior.
Additionally, we proposed a quantitative comparison by studying CNNs illusions through asymmetric color matching experiences previously done by humans~\cite{ware82}.
In those experiments we explored simple CNN architectures (with 2 or 4 hidden layers) trained for image denoising, deblurring and restoration.
And we also studied the behavior of a recent, much deeper CNN trained for the same kind of task: the 17 layers architecture of Zhang et al.~\cite{zhang2017beyond} pretrained for denoising.

Qualitative analysis shows that simpler networks do modify their response in the same
direction as the humans in most cases. However, the 17-layer network leads to negligible illusions (negligible shifts in the response).
On the other hand, quantitative results on asymmetric matching (simple center-surround setting) show that illusions in the networks
with 4-hidden layers are the same order of magnitude than those of humans. However, the perceived colors suffer from assimilation rather
than contrast (as opposed to humans). Assimilation in center-surround settings is also the behavior of the 2-hidden layer networks
although with a weaker illusion. Finally, the 17-layer network displays almost negligible illusion in the corresponding pair experiment,
consistently with the negligible shifts found in the response.
Therefore, the quantitative analysis reveals that simple networks in center-surround settings have substantial illusions,
but not human-like.

The proposed eigenanalysis of simple networks reveals interesting similarities with human vision and suggests a reason for the assimilation
behavior found in simple center-surround settings.

Using mild assumptions, we found that these simple networks \emph{implicitly} operate in an opponent color space,
and in this human-like color representation, they spatially filter the signal with markedly different bandwidths in the
achromatic channel and the red/green and yellow/blue channels. The filters found resemble the human contrast sensitivities
in opponent channels.

This simple linear description, which explains about 95\% of the response of the simple networks (but arguably a smaller proportion of human behavior)
may explain the assimilation illusion in the networks: in center-surround settings the low-pass nature of the filtering in the chromatic channels
shifts the hue of the target towards the hue of the surround.
\vspace{0.2cm}

From the results and associated analysis, these considerations may follow.

\paragraph{Visual illusions and image statistics} The findings in this work are consistent with the long-standing tradition in vision science that considers low-level visual illusions as by-product of optimization of visual systems to perform basic tasks in natural environments~\cite{Barlow90,Clifford00,Clifford02,Cliford07}.
More specifically, the results of our linear analysis link the behavior of the networks with (a) the statistical basis of the image restoration problem
and, more interestingly, with (b) optimal coding theories of human vision.
First, it is interesting to note that our linear analysis of the networks leads to functions that resemble the principal directions of natural colors~\cite{Buchsbaum83,sejnowski2002}, and natural color images~\cite{Sejnowski2001,Hyvarinen09,Gutmann14}.
However, note that in our analysis we are not diagonalizing the covariance matrix of natural signals.
In fact, signal decorrelation or independence was not enforced in anyway.
Therefore, although similar, our result should not be attributed to information maximization or sparse coding~\cite{Olshausen1996,Sejnowski2001}.
Instead, as stated above, if the goal is removing non-natural features from stimuli, Wiener or Tikhonov ideas to focus in the places where the signal is relevant~\cite{wiener1949extrapolation,tikhonov1977solutions}, naturally lead to filters matched to the signal spectrum.
With this in mind, and given the relation between the average spectrum of the signals, their autocorrelation and their covariance,
it makes sense that the optimal filter obtained from our linear approximation is very similar to the covariance of the natural signals.
Therefore, our result to reconstruct signals with minimum error turns out to be very similar to the PCA result.
Nevertheless, the reason why CNNs trained for image restoration develop opponent chromatic channels and CSF-like filters would be more in line with signal/noise explanations of visual function~\cite{Atick92}. Note that error minimization and information maximization are similar, but not the same (see ~\cite{Lloyd82leastsquares} for the original account, and see~\cite{MacLeod01,MacLeod03,Laparra12,Laparra15} for sequels in vision science).

\paragraph{The limitations of ANNs to study vision}
Our psychophysical-like analysis of ANNs shows that while they are deceived by illusions, their response might be significantly different to that of humans: these discrepancies with humans in quantitative experiments imply a word of caution on using ANNs to study human vision. As mentioned earlier, ANNs were inspired by classical biological models of vision, and for this reason they share the L+NL formulation \cite{Haykin2009} of the ``standard'' model of vision \cite{Olshausen2005}.  But this model is questioned in the vision science literature.
 Vision models and ANNs
use L+NL modules derived from fitting some data, and in every case either the linear filters are constant or the models do not have general rules as to how the filters should be modified depending on the input \cite{Betz2015,Li2019}.
This is an essential weakness of all these models,
because
visual adaptation,
an essential feature of the neural systems of all species, produces
a change in the input-output relation of the system that is driven by the stimuli \cite{Wark2009},
and therefore
requires that the linear and/or the nonlinear stages of a L+NL model change with the input in order to explain neural responses \cite{Meister1999,Coen2012,Jansen2018}.
L+NL models are not tests of how well the linear filter of a neuron describes its behavior, they have been obtained simply by {\it assuming} that the neuron performs a linear summation and then searching for the best-fitting linear model.
In neuroscience, the standard model explains at the most a $40\%$ of the data variance in V1, and the performance decays substantially when the input stimuli are natural images instead of the usual synthetic inputs \cite{Olshausen2005}, suggesting that a more complex, network nonlinearity is at work \cite{Carandini2005}.
In many situations, experimental data on visual perception contradicts the central notions of L+NL models \cite{Wandell1995}.
It has been proposed \cite{Olshausen2013} that the standard model is not just in need of revision, it is the wrong starting point and needs to be discarded altogether.

\section*{Acknowledgements}

This work has received funding from the European Union’s Horizon 2020 research and innovation programme under grant agreement number 761544 (project HDR4EU) and under grant agreement number 780470 (project SAUCE), and by the Spanish government and FEDER Fund, grant ref. PGC2018-099651-B-I00 (MCIU/AEI/FEDER, UE). JM has been supported by the Spanish government under the MINECO grant ref. DpI2017-89867 and by the Generalitat Velanciana grant ref. grisoliaP-2019-035. We gratefully acknowledge the support of NVIDIA Corporation with the donation of the Titan Xp GPU used for this research.

\bibliographystyle{elsarticle-num}
\bibliography{egbib}







\end{multicols}

\end{document}